\documentclass{sig-alternate}
\usepackage{color}
\usepackage{overpic}
\usepackage{subfigure}
\usepackage{algorithmic}
\usepackage[ruled]{algorithm2e}

\begin{document}

\title{Photo Stylistic Brush: Robust Style Transfer via Superpixel-Based Bipartite Graph}
\numberofauthors{4} 
\author{
\alignauthor
Jiaying Liu \\
       \affaddr{Peking University}\\
       \email{liujiaying@pku.edu.cn}
\alignauthor
Wenhan Yang \\
       \affaddr{Peking University}\\
       \email{yangwenhan@pku.edu.cn}
\and
\alignauthor 
Xiaoyan Sun \\
       \affaddr{Microsoft Research Asia}\\
       \email{xysun@microsoft.com}
\alignauthor 
Wenjun Zeng \\
       \affaddr{Microsoft Research Asia}\\
       \email{wezeng@microsoft.com}
}

\maketitle
\begin{abstract}
With the rapid development of social network and multimedia technology, customized image and video stylization has been widely used for various social-media applications. In this paper, we explore the problem of exemplar-based photo style transfer, which provides a flexible and convenient way to invoke fantastic visual impression. Rather than investigating some fixed artistic patterns to represent certain styles as was done in some previous works, our work emphasizes styles related to a series of visual effects in the photograph, \emph{e.g.} color, tone, and contrast. We propose a photo stylistic brush, an automatic robust style transfer approach based on \textbf{Superpixel}-based \textbf{BI}partite \textbf{G}raph~(\textbf{SuperBIG}). A two-step bipartite graph algorithm with different granularity levels is employed to aggregate pixels into superpixels and find their correspondences. In the first step, with the extracted hierarchical features, a bipartite graph is constructed to describe the content similarity for pixel partition to produce superpixels. In the second step, superpixels in the input/reference image are rematched to form a new superpixel-based bipartite graph, and superpixel-level correspondences are generated by a bipartite matching. Finally, the refined correspondence guides SuperBIG to perform the transformation in a decorrelated color space. Extensive experimental results demonstrate the effectiveness and robustness of the proposed method for transferring various styles of exemplar images, even for some challenging cases, such as night images.
\end{abstract}

\vspace{-2mm}

\section{Introduction}
\label{sec:intro}
\begin{figure}[!ht]
	\centering\includegraphics[width=8.5cm]{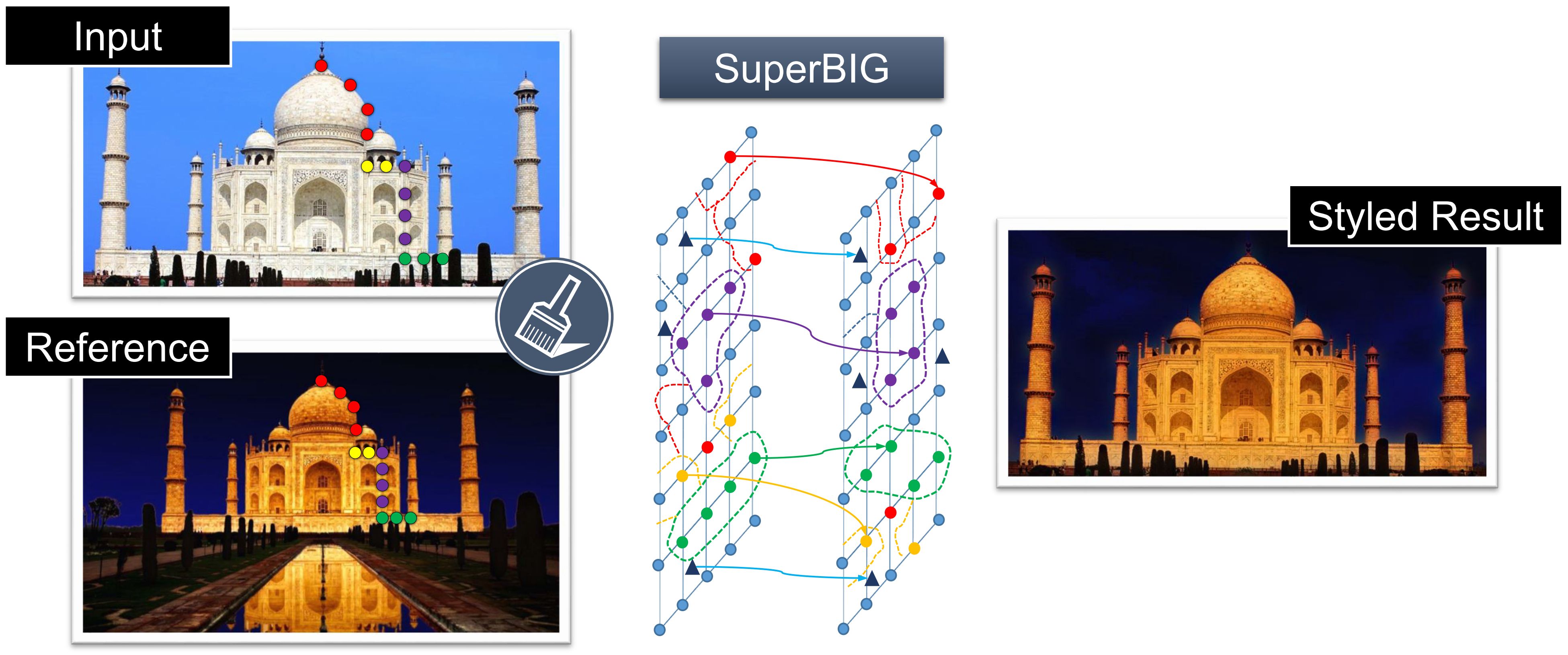}
    \vspace{-5mm}
	\caption{Illustration of the proposed stylistic brush, SuperBIG.}
	\label{fig:two_matchings}
\end{figure}
With the prevalence of multimedia social networking, it has become popular to share photos online. Most people nowadays prefer uploading photos with special artistic enhancement made by various Apps such as Facebook and Instagram instead of the original ones. This kind of photo style enhancement makes pictures dramatically more impressive and inspires new imagination. However, existing systems either allow users to only roughly change the photo in a fixed template, or require a series of subtle processes by experienced photographers using the editing software.

Image style transfer aims to automatically change the \emph{stylistic elements} of an input image (color, texture, contrast, \emph{etc.}) to follow a given exemplar, \emph{e.g.} well-known paintings or fabulous pictures taken by professional photographers. Early works start by transferring one of these elements among images. The color transfer methods either extract the most representative colors from the images and build a conversion algorithm between those colors~\cite{reinhard2001color,wang2010data}, or directly adjust the color distribution via a histogram feature fitting~\cite{pouli2010progressive,pitie2007automated}. Contrast is usually transferred in the frequency band space, such as the bilateral space~\cite{bae2006two}, Laplacian pyramid~\cite{li2005compressing} or Haar pyramid~\cite{sunkavalli2010multi}. Since these methods only consider one specific stylized element, they may produce some visual effect, but are difficult to be applied widely in practice.

Meanwhile, the image stylization is also explored in the computer graphics community, referred to as non-photorealistic rendering~(NPR). It aims to generate non-photorealistic style images, such as watercolor painting~\cite{hertzmann2003survey}, sketch generation~\cite{wang2009face} and abstract drawing~\cite{curtis1997computer}. By a carefully crafted design, a bunch of stylized elements are extracted to represent the artistic style of an image and further used to transfer artistic visual effects. However, these hand-crafted features, designed with certain type of artworks, lack expandability by nature and are not adaptive in representing other styles or new styles.

In real applications, it is unrealistic to ask most people to give a specific description about what style they exactly want. Usually, what they could offer is a real example they saw before, \emph{e.g.} `Mona Lisa', or an abstract word they read from books, \emph{e.g.} `Baroque'. Knowing little about image editing, they need a tool to define a bunch of style settings from these examples and make adjustments automatically. Like the \emph{format painter} of Microsoft Office, \emph{Stylistic Brush} provides a desirable and powerful tool to enable an automatic arbitrary style transfer between images. More specifically, this functionality could be implemented by exemplar-based stylization, as shown in Figure \ref{fig:two_matchings}. The style is extracted dynamically from the fantasy \textit{reference image}~(also referred to as target image). A new output image is synthesized based on the content of the input image and the extracted styles of the reference one.

Therefore, some works investigating image stylization by considering the style composition instead of a single style element are emerging. Most of these methods devote to separating and dealing with the content and style individually. An early work~\cite{hertzmann2001image} explored the concept of `image analogy' by building a multiscale autoregression framework to adaptively learn a wide variety of ``image filter''. Zhang \textit{et al.}~\cite{zhang2013style} proposed to perform an image component analysis to decompose an image into three components and constructed a coarse-to-fine Markov random field to propagate colors in the paint and edge components. In~\cite{NNart}, a deep network-based method was proposed to separate and recombine the content and style. A composition of the learned CNN features gives a clue of content correspondence and guides the production of new artistic images via transferring the style features.

These methods suffer from two limitations: 1) From the model aspect, the assumption that the content and the style could be separable may be questional. Some common observations, such as sunset with red color and grass with certain texture patterns, lead to the conclusion that some styles are highly correlated with the image content. Thus, previous methods with such a separable assumption lose some style information in the transformation. 2) From the application aspect, these methods mainly focus on painting styles and are good at transferring or generating texture styles. However, in real applications of photography, people usually pay more attention to visual effects caused by color, light, contrast~\textit{etc.} than textures.

In this paper, we aim to create a stylistic brush to help people beautify their photos by transferring desirable styles of a chosen exemplar image to the input one. Focusing on photos, we pay more attention to the color, light and contrast of a photograph instead of the factors related to art, such as textures or strokes. Compared to previous methods, we make two more reliable assumptions: 1) For most photos, the Internet enables us to collect a content similar reference with a favorable style. It is usually the case for a certain category of images, such as the landmark or face images; 2) Different from general content-based features, we obtain matched points of the same scene between the reference and input images as more reliable guidance of content similarity, via dense correspondence detection methods.

With the above considerations, the proposed stylistic brush is realized by a robust style transfer method based on the Superpixel BIpartite Graph~(SuperBIG) framework for image stylization. First, a dense correspondence between the input and reference images is estimated to obtain matched pixels as the primitives. By exploiting hierarchical features in different-granularity, we measure the distances from pixels to the identified matched points in the feature space to cluster these pixels into superpixels. Then a bipartite graph partition is exploited to assign unclustered pixels into superpixels by considering both the local and global consistency. Afterwards, superpixels of two images are rematched to form a new superpixel bipartite graph to refine the final superpixel-level correspondent relationship. Finally, SuperBIG transfers colors within each superpixel correspondence in a decorrelated color space to achieve the stylization.

The main contributions of our work are summarized as follows:
\begin{itemize}
	\item We analyze the challenges for practical photo stylization, and propose ``Stylistic Brush'' to solve this problem integrally, \emph{i.e.} stylizing the input image based on the styles of a given exemplar. To the best of our knowledge, this is the first attempt to transfer complex natural photo styles instead of painting strokes by example images.

	\item We propose an automatic robust style transfer framework based on the Superpixel BIpartite Graph~(SuperBIG). It estimates the superpixels from the input image and performs correspondence matching between these superpixels jointly by a two-step bipartite partition and matching. This step-by-step abstraction integrates the local consistency of superpixel and the global matching of the bipartite graph effectively.

	\item Benefiting from diversity of the proposed hierarchical features in different granularity, as well as the advantages of the unified bipartite graph framework, SuperBIG achieves promising results in terms of effectiveness and robustness in extensive experiments, even for some challenging cases, such as night images.
\end{itemize}

\section{Related Works}
\label{sec:related}
\subsection{Non-Photorealistic Rendering}
Non-photorealistic rendering was first proposed by Winkenbach and Salesin \cite{winkenbach1994computer}. It aims to produce images derived from a wide variety of styles such as painting, drawing, sketching, illustration and animation for digital art. Non-experts can transfer artistic styles of famous painters to ordinary photos taken everyday with the help of NPR. Nowadays, many ad-hoc NPR schemes have been proposed for this task with a varying degree of success \cite{kyprianidis2013state}. While Li \emph{et al.} \cite{li2008automated} proposed to create and view interactive exploded views of 3D models, Pouli and Reinhard \cite{pouli2011progressive} utilized a user-specified target image's color palette to achieve creative effects. For artistic styles rendering, some researchers focus on simulating virtual brush strokes to obtain a particular style \cite{hertzmann2001paint,lin2010painterly}. Region-based methods are also used to independently render the interiors of regions \cite{gooch2002artistic,wang2010video}. In the meantime, many image processing filters have been applied to produce images in artistic styles \cite{orzan2007structure,kyprianidis2009image}. Different from NPR studying on artistic patterns rendering, our work aims to address the challenge of photo style transfer, where more diversified styles are faced and photometric properties, such as light, contrast, change more abruptly within an image.
\vspace{2mm}

\subsection{Hand-Crafted Style Transfer}
Hand-crafted style transfer techniques aim to adjust the color, contrast and tone of images, with the aid of signal properties, \emph{e.g.} the statistic information of colors, without considering the content-level correspondence. For color transfer, the work in~\cite{reinhard2001color} transferred colors by matching the statistics of color distributions. Subsequent works improved the accuracy and robustness of statistical estimation, such as soft-segmentation~\cite{tai2007soft}, multi-dimensional distribution matching~\cite{pitie2005n} and minimal displacement mapping~\cite{pouli2011progressive}. There are also some methods~\cite{huang2009landmark,hwang2014color} that consider colorizing the image with user defined colors. These methods propagate colors with an elaborately designed constraint to ensure natural visual effect of the produced result. For contrast and tone, adjustment is manipulated in the frequency domain, such as bilateral space~\cite{bae2006two}, Laplacian pyramid~\cite{li2005compressing} or Haar pyramid~\cite{sunkavalli2010multi}. Our work focuses on transferring photo styles adaptively based on the given references instead of a crafted architecture designed for the transfer of a certain style.
\begin{figure*}[!ht]
	\centering\includegraphics[width=16.5cm]{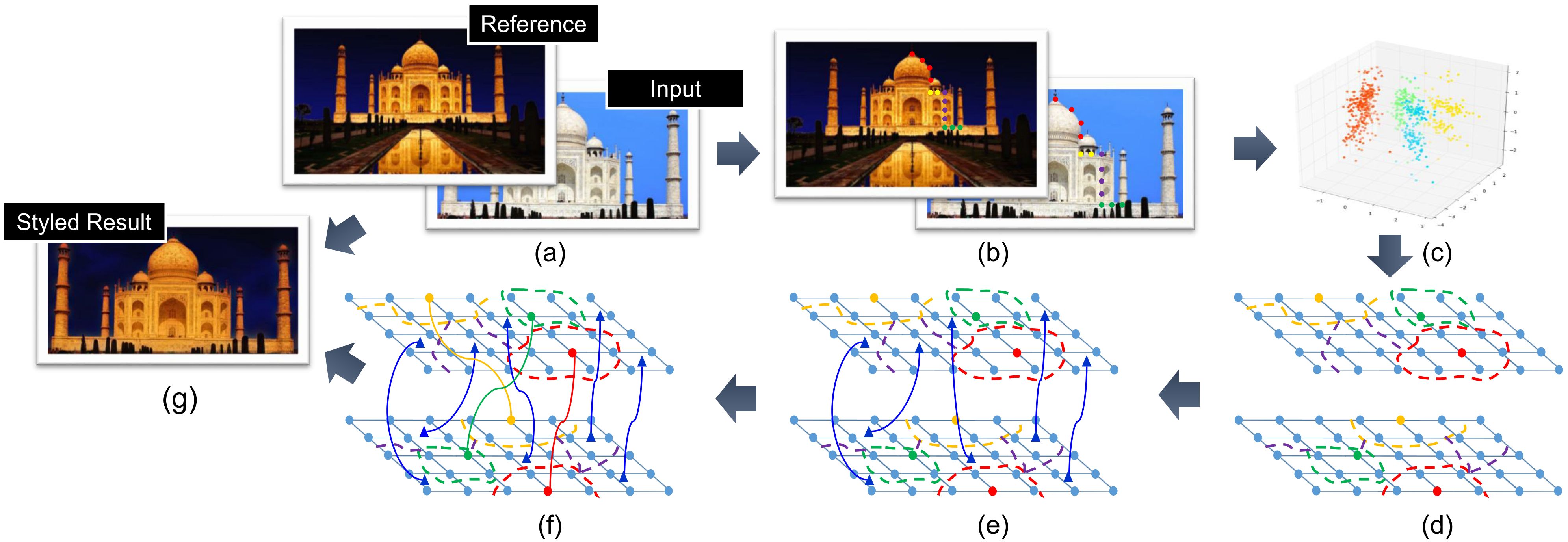}
	\vspace{-4mm}
	\caption{The flowchart of SuperBIG algorithm. (a) Input and reference images. (b) Matched points detected by dense correspondence method. (c) Hierarchical features for each pixel. (d) Superpixels obtained by the distance between each pixel and matched points. (e) Superpixels obtained by pixel-level bipartite graph partition. (f) The superpixel correspondence generated by superpixel bipartite graph matching. (g) The styled result based on colors of input and reference images, as well as the superpixel correspondence.}
	\label{fig:flowchart}
\end{figure*}

\subsection{Example-Based Style Transfer}
For image stylization, only exploiting signal properties and statistical correspondence cannot guarantee the correctness of the local style decision. Recently, some methods explore ways to create and utilize the content-level correspondence to benefit the stylization. In~\cite{irony2005colorization,levin2004colorization}, the input and reference images are segmented first. Then, colors are propagated from color images to greyscale images via a set of locally homogeneous patches or basic elements called color scribbles. Charpiat \textit{et al.}~\cite{huang2005adaptive} assigned colors to the greyscale image by solving an optimization problem in the framework of graph cut. In~\cite{chia2011semantic}, after manual segmentation of major foreground objects, a belief-propagation colorizes the greyscale image with the help of Internet images. In~\cite{chia2011semantic,liu-2008-intrinsic}, colors are transferred by estimating per-pixel registered correspondence between input and reference images. Kumar \textit{et al.}~\cite{gupta2012image} proposed to create correspondences between superpixels by fast cascade feature matching, and then refine the transfer results by a voting approach. Cheng \emph{et al.}~\cite{cheng2015color} proposed a superpixel-based recoloring scheme based on a soft matching embedded with color statistics, texture characteristics and spatial constraints to generate new recolored images. There are also some works that aim to conduct favorite exemplars recommendation based on visual information~\cite{cheng2015color} or patch aggregation~\cite{lu2015deep}. Compared with previous methods, our method aims to address the general style transfer of photos instead of a certain style element, such as only color, or some artistic styles. We devote to offering an integrated solution to transfer the composition of light, color, contrast automatically.

\section{Superpixel Bipartite Graph for Photo Style Transfer}
\label{sec:proposed}
The proposed SuperBIG transfers the style of the reference image to the input image by a two-step bipartite graph framework as shown in Figure~\ref{fig:flowchart}. SuperBIG first detects the dense correspondence~(Figure~\ref{fig:flowchart}(b)) and calculates the designed hierarchical features~(Figure~\ref{fig:flowchart}(c)). Based on the correspondence and features, SuperBIG then aggregates pixels into superpixels using a simple clustering algorithm~(Figure~\ref{fig:flowchart}(d)) for the pixels around the matched points and a bipartite graph framework~(Figure~\ref{fig:flowchart}(e)) for the pixels far from the matched points. Afterwards, SuperBIG transfers the colors between corresponding superpixels~(Figure~\ref{fig:flowchart}(f)) in a decorrelated color space.

\subsection{Superpixel Aggregation with Hierarchical Features}
Superpixel is a pixel cluster consisting of several pixels with similar color and brightness. It is proposed to well define coherent regions, as basic elements of over-segmentation. It usually provides an initialization for segmentation~\cite{yang2008unsupervised,wang2008normalized,mobahi2011segmentation} or a soft constraint on segmentation~\cite{fern2004solving,li2012segmentation}. Compared with raw pixels, superpixel is a more sparse and efficient representation, while it provides more reliable and fine-grained regions in comparison with segmented objects.

SuperBIG creates and embeds superpixels of input and reference images in a unified bipartite graph framework. It obtains superpixels through two steps. The first one is to cluster pixels into superpixels based on distance measurement with dense correspondence, which is estimated by deep matching~\cite{DeepMatch}. The relevant hierarchical features for measuring the distances between pixels include colors, intensity patterns, textures, \emph{etc.} The second step is to employ an automatic bipartite partition in a unsupervised way to group pixels that are not covered by any superpixel in the first step. Here we elaborate on the related features.

We use the subscript $(i,j)$ to index the pixel location of an image $\mathbf{I}$ and utilize superscript $c$ and $f$ to denote features of the input and reference images, respectively. $\mathbf{I}_{(i,j)}$ is defined as the intensity of a pixel at the location $(i,j)$. We extract a set of features for the following two purposes: To measure the content similarity in the same domain/style (\emph{e.g.} within an image) or to measure that cross domains/styles (\emph{e.g.} in two styled images). Thus, the extracted features are classified into two categories: style-related (including patch intensity, color, gradient, absolute location) and style-independent (including texture, relative location, locality-constrained linear coding feature). All these extracted features are described below,

\begin{itemize}
	\item Intensity vector of a patch:
		\begin{equation}
            {\mathbf{M}_{(i,j)} = \left[\mathbf{I}_{(k,l)}\right]^T\Big|_{(k,l)\in \mathcal{N}_{(i,j)}}}, \nonumber
		\end{equation}
		where the set $\mathcal{N}_{(i,j)}$ contains locations of pixels $(k,l)$ in a patch centered at the location $(i,j)$.

	\item Color $\mathbf{C}_{(i,j)}$ at pixel $(i,j)$, which is composed of,
	\begin{equation}\mathbf{C}_{(i,j)} = \left[\mathbf{I}_{\mathbf{R}(i,j)},\mathbf{I}_{\mathbf{G}(i,j)},\mathbf{I}_{\mathbf{B}(i,j)}\right]^T, \nonumber
	\end{equation}
	where $\mathbf{I}_{\mathbf{R}}$, $\mathbf{I}_{\mathbf{G}}$ and $\mathbf{I}_{\mathbf{B}}$ are three channels of an image~$\mathbf{I}$. They are related to the intensity of that pixel as follows,
	\begin{equation}\mathbf{I}_{(i,j)} = \sqrt{ \mathbf{I}_{\mathbf{R}(i,j)}^2+\mathbf{I}_{\mathbf{G}(i,j)}^2+\mathbf{I}_{\mathbf{B}(i,j)}^2}. \nonumber
	\end{equation}

\item Gradient of a patch:
    \begin{equation}
    \mathbf{DV}_{(i,j)} = \left[\left\{\sqrt{{\mathbf{d}\mathbf{I}_{{x(k,l)}}^2+\mathbf{d}\mathbf{I}_{{y(k,l)}}^2}}\Big|{(k,l)\in \mathcal{N}_{(i,j)}}\right\}\right]^T,\nonumber
    \end{equation}
where $\mathbf{d}\mathbf{I}_{{x}}$ and $\mathbf{d}\mathbf{I}_{{y}}$ denote the intensity variation of the original image along horizontal and vertical directions, respectively.

\item Absolute location:
    \begin{equation}
    \mathbf{L}_{(i,j)}^a = \left[\frac{i}{h},\frac{j}{w}\right]^T,\nonumber
    \end{equation}
where $h$ and $w$ are the height and width of an image. It is defined as the normalized location in the original coordinates for the image.

\item Texture feature $\mathbf{T}_{(i,j)}$ of a patch centered at pixel $(i,j)$. Details about its calculation are presented in \cite{FBTS_2015_JY_TIP}.

\item Relative location, $\mathbf{L}_{(i,j)}^r$. SuperBIG regards the dense points as reliable locations and utilizes them to `relocate' the pixels with the novel coordinates, which takes the locations of these matched points as the basis. It is defined as the representation coefficients of a pixel location, when taking locations of several nearest matched points within the image as the basis. Locations of five nearest matched points to pixel $(i,j)$ are denoted as,
	\begin{equation}
	\pmb{\tau} = \left\{\left[i_{l},j_{l}\right]^T|_{l=1,2,...,5}\right\}.\nonumber
	\end{equation}
	The current location $(i,j)$ is represented by the multiplication of $\pmb{\tau}$ and a representation coefficient $\pmb{\alpha}$,	
	\begin{equation}
	\pmb{\tau}\pmb{\alpha}=\left[i,j\right]^T.\nonumber
	\end{equation}
	Then, $\pmb{\alpha}$ is solved by,
	\begin{equation}
	\pmb{\alpha}=(\pmb{\tau}^T\pmb{\tau}I+n_{\pmb{\alpha}})^{-1}(\pmb{\tau}^T\left[i,j\right]^T),\nonumber
	\end{equation}
where $n_{\pmb{\alpha}}$ is the ridge parameter for $\pmb{\alpha}$ to avoid singular solutions. To generate the relative location $\mathbf{L}_{(i,j)}^r$, we put the solved $\pmb{\alpha}$ to $\mathbf{L}_{(i,j)}^r$ in the corresponding dimension that belongs to the matched point and zeros in other dimensions.

\item Locality-constrained linear coding~(LLC) feature, $\mathbf{S}_{(i,j)}$. Similar to the idea of calculating the relative location, we calculate the `relative location' in the feature space, to generate a measurement of content similarity, independent on the style. Similarly, with the matched points provided by deep matching, we use features of these matched points as the basis (or the coordinates in the feature space) to calculate the representation coefficients, independent on the style.
	Assume the five nearest matched points at the location $(i,j)$ are represented in the feature space,
	\begin{equation}
	\pmb{\tau}_f = \left\{\left[\mathbf{M}_{i_l,j_l},\mathbf{C}_{i_l,j_l},\mathbf{I}_{i_l,j_l},\mathbf{DV}_{i_l,j_l} \right]^T|_{l=1,2,...,5}\right\}.\nonumber
	\end{equation}
	Then, a sparse coefficient $\pmb{\beta}$ is calculated by solving,
	\begin{equation}
	\pmb{\tau}_f\pmb{\beta}=\left[\mathbf{M}_{i,j},\mathbf{C}_{i,j},\mathbf{I}_{i,j},\mathbf{DV}_{i,j} \right]^T.\nonumber
	\end{equation}
	We then have,
	\begin{equation}	\pmb{\beta}=(\pmb{\tau}_f^T\pmb{\tau}_f+n_{\pmb{\beta}}I)^{-1}(\pmb{\tau}_f^T\left[\mathbf{M}_{i,j},\mathbf{C}_{i,j},\mathbf{I}_{i,j},\mathbf{DV}_{i,j} \right]^T),\nonumber
	\end{equation}
where $n_{\pmb{\beta}}$ is the ridge parameter for $\pmb{\beta}$ to avoid singular solutions. To generate the relative location $\mathbf{S}_{(i,j)}^r$, we put the solved $\pmb{\beta}$ to $\mathbf{S}_{(i,j)}^r$ in the corresponding dimension that belongs to the matched point and zeros in other dimensions.

With the help of the above mentioned features of several nearest matched points $\mathbf{P}_{(i,j)}^c$ or $\mathbf{P}_{(i,j)}^f$, $\mathbf{S}_{(i,j)}^c$ and $\mathbf{S}_{(i,j)}^f$ are representation coefficients of the unmatched points $(i,j)$ from the input and reference images.
\end{itemize}

Intuitively, these features are diverse in order to cover most information to build the content correspondence. As mentioned above, according to whether a feature is capable of measuring the content similarity cross styles, these features are classified into: style-related and style-independent. The former is mainly utilized to measure the similarity between input and reference images, while the latter is exploited to measure the similarity between two pixels in the same image.

Here we create superpixels around matched points and build a mapping based on the correspondences of these points. Intuitively, coupled superpixels around paired matched points share the same style transformation. We use $p$ and $q$ to index two arbitrary pixels in the input and reference images, respectively. And let $t$ index an arbitrary pixel in one of them. For each pair of matched point locations $(i_p, j_p)$ and $(i_q, j_q)$, the distance of one pixel $(i_t, j_t)$ in the input image to the corresponding matched point in the reference image is calculated by style-dependent features as follows,
\begin{eqnarray}
\label{equ:distance2}
&& \mathbf{D}^c(u_{(i_t, j_t)},u_{(i_p, j_p)})\nonumber \\&=& -\frac{\left\|\mathbf{M}^{c}_{(i_t,j_t)}-\mathbf{M}^{c}_{(i_p, j_p)}\right\|_2^2}{\lambda_\mathbf{\mathbf{M}}}-\frac{\left\|\mathbf{T}^c_{(i_t,j_t)}-\mathbf{T}^c_{(i_p, j_p)}\right\|_2^2}{\lambda_\mathbf{T}}  \nonumber \\
&&-\frac{\left\|\mathbf{C}^{c}_{(i_t,j_t)}-\mathbf{C}^{c}_{(i_p, j_p)}\right\|_2^2}{\lambda_\mathbf{\mathbf{C}}}-\frac{\left\|\mathbf{DV}^{c}_{(i_t,j_t)}-\mathbf{DV}^{c}_{(i_p, j_p)}\right\|_2^2}{\lambda_\mathbf{\mathbf{DV}}}	\nonumber \\
&&-\frac{\left\|\mathbf{L}^{a,c}_{(i_t,j_t)}-\mathbf{L}^{a,c}_{(i_p, j_p)}\right\|_2^2}{\lambda_\mathbf{\mathbf{L}^{a}}},
\end{eqnarray}
where $\lambda_{(\cdot)}$ are weighting parameters to balance the effect of each term. The distance $\mathbf{D}^f(v_{(i_t, j_t)},v_{(i_q, j_q)})$ in $\mathbf{I}^f$ can be computed similarly. Then, we create super-pixel clusters $\mathbf{F}^{c,m}_p$ and $\mathbf{F}^{r,m}_q$ containing all the pixels with a distance to $p$ and $q$ respectively less than a given threshold $\mathcal{T}_{cluster}$. After that, superpixels around the matched points are obtained. SuperBIG further deals with other unsettled pixels in a bipartite graph framework hereafter.

\subsection{Pixel Bipartite Graph Partition}
\label{sec:WBGM}
After obtaining the superpixel around matched points, SuperBIG constructs a pixel-level bipartite graph from the uncovered pixels that do not belong to any given superpixel. Afterward, a bipartite partition is followed to cluster those unsettled pixels into superpixels.

Let $\mathbf{f}_{(i,j)}^c$ and $\mathbf{f}_{(i,j)}^r$ represent the hierarchical features corresponding to the pixel located at $(i,j)$ in the input and reference images. Because we aim to calculate the content closeness of pixels in two images with different styles, the hierarchical features consist of style-free features, such as locations, gradient, textures, defined as follows, \vspace{-1mm}
\begin{equation}
\label{equ:hifeature}
 \mathbf{f}_{(i,j)}^c =  \left[ \mathbf{S}_{(i,j)}^c, \mathbf{T}_{(i,j)}^c, \mathbf{L}_{(i,j)}^{a,c}, \mathbf{L}_{(i,j)}^{r,c}\right].
\end{equation}
So does $\mathbf{f}_{(i,j)}^r$.

Based on the hierarchical features to calculate the affinities between nodes, SuperBIG constructs the pixel bipartite graph. Let $u_{(i,j)}$ and $v_{(i,j)}$ denote the node corresponding to the pixel in the location $(i,j)$ of the input and reference image, respectively. Here $(i,j)$ only represents the location of unsettled pixels. There is an edge connection between corresponding nodes in the bipartite graph, only when the nearest dense points of their corresponding pixels are largely matched. Then, the pixel corresponds to the node in the graph, and edge weights (affinities) are calculated based on hierarchical features $\mathbf{f}_{(i_p,j_p)}^c$ and $\mathbf{f}_{(i_q,j_q)}^f$ adjusted by weighting parameters $\lambda_{(\cdot)}$ for each kind of features as follows,
\begin{eqnarray}
	\label{equ:distance}
	&& \mathbf{E}(u_{(i_p,j_p)},v_{(i_q,j_q)}) \nonumber \\
    &&=\exp{\left\{-\frac{\left\|\mathbf{S}^{c}_{(i_p,j_p)}-\mathbf{S}^{f}_{(i_q,j_q)}\right\|_2^2}{\lambda_\mathbf{\mathbf{S}}}-\frac{\left\|\mathbf{T}^c_{(i_p,j_p)}-\mathbf{T}^f_{(i_q,j_q)}\right\|_2^2}{\lambda_\mathbf{T}}\right.}  \nonumber \\	&&-\left.\frac{\left\|\mathbf{L}^{a,c}_{(i_p,j_p)}-\mathbf{L}^{a,f}_{(i_q,j_q)}\right\|_2^2}{\lambda_\mathbf{\mathbf{L}^{a}}}-\frac{\left\|\mathbf{L}^{r,c}_{(i_p,j_p)}-\mathbf{L}^{r,f}_{(i_q,j_q)}\right\|_2^2}{\lambda_\mathbf{\mathbf{L}^{r}}}\right\}.
\end{eqnarray}

Then, a weighted bipartite graph is constructed between two nodes $(u,v)$, corresponding to the pixels of images that are exactly paired matched points in the dense correspondence. Their edge weights (affinities) $\mathbf{E}(u,v)$ correspond to the similarities, which are independent of the style.

When performing the graph partition, a natural choice is spectral clustering. It is exploited to capture the cluster structure of a graph by clustering the spectrum of the Laplacian matrix. $\mathbf{D}$ is defined as the degree matrix. It is formulated as a generalized eigen-problem,
\begin{eqnarray}
\label{equ:par1}
\mathbf{J}\mathbf{g} = \lambda \mathbf{D}\mathbf{g},
\end{eqnarray}
where $\lambda$ is the eigenvalue to be optimized. And $\mathbf{J}= \mathbf{D}-\mathbf{\Omega}$ is the Laplacian matrix and $\mathbf{D} = diag(\mathbf{\Omega1})$ is the degree matrix. $\mathbf{1}$ is a unit vector and $\mathbf{\Omega}$ denotes the affinity (adjacent) matrix of the graph, that contains the affinity $\mathbf{E}(u,v)$ of every paired nodes $(u,v)$ in the graph. For clustering, the Laplacian matrix is approximated by a block-diagonal matrix including $k$ eigenvalues block-diagonal matrix. The Laplacian matrix can be also defined as the normalized Laplacian $\mathbf{J}_N = \mathbf{D}^{{-1}/{2}}\mathbf{J}\mathbf{D}^{{-1}/{2}}$ or generalized Laplacian $\mathbf{J}_G = \mathbf{D}^{{-1}}\mathbf{J}$.

It can be solved with the Lanczos method~\cite{golub2012matrix} on the normalized affinity matrix $\tilde{\mathbf{\Omega}}=\mathbf{D}^{-1/2}\mathbf{\Omega}\mathbf{D}^{1/2}$ or partial SVD~\cite{zha2001bipartite} on normalized across-affinity matrix. Adopting the latter solution in our method, the bottom $k$ eigenvectors of (\ref{equ:par1}) are obtained by the top $k$ left and right singular vectors of the normalized across-affinity matrix,
    \begin{eqnarray}
    \label{equ:par2}
    \tilde{\mathbf{\Omega}}_a = \mathbf{D}^{-1/2}_{X}\mathbf{\Omega}\mathbf{D}^{-1/2}_{Y},
    \end{eqnarray}
where $\mathbf{D}_X=diag(\mathbf{\Omega})\mathbf{1}$ and $\mathbf{D}_Y=diag(\mathbf{\Omega})^T\mathbf{1}$ denote the degree matrix of $\mathbf{X}$ and $\mathbf{Y}$, respectively. Then, we obtain $k$ superpixel clusters $\mathbf{F}^{c,u}_{p}$ and $\mathbf{F}^{r,u}_{q}$ and get a set of coupled superpixel clusters $\mathbf{F}^{c} = \left[\mathbf{F}^{c,m},\mathbf{F}^{c,u}\right]$ and $\mathbf{F}^{r} = \left[\mathbf{F}^{r,m},\mathbf{F}^{r,u}\right]$.

\subsection{Superpixel Bipartite Graph Matching}
In the above step, SuperBIG estimates the superpixels for the pixels that are not covered by superpixels of matched points. In this process, superpixels of matched points and their covered pixels are totally ignored in the constructed pixel-level bipartite graph. It may lead to inaccurate matchings when some superpixels of matched pixels in the input image in fact correspond to the superpixels of unmatched pixels in the reference image.

Thus, SuperBIG constructs a superpixel bipartite graph and performs a graph matching on it. The nodes of the new graph represent superpixels of $\mathbf{F}^{c}$ and $\mathbf{F}^{r}$. There is an edge connection between corresponding nodes, only when their hierarchical features are close enough in the feature space. Considering that the pixels in a superpixel share similar features, for similarity, hierarchical features of a superpixel are defined as the mean vector of hierarchical features of pixels within it. And the affinities between superpixel bipartite graph are calculated based on the superpixel hierarchical feature, in the same way as (\ref{equ:distance}). Then, SuperBIG solves the bipartite graph matching by the Hungarian algorithm~\cite{bruff2005assignment}, obtaining final superpixel correspondences $\mathbf{F}^{c}_f$ and $\mathbf{F}^{r}_f$.

\subsection{De-Correlated Style Transfer}
\label{sec:LCP}
After we obtain a reliable superpixel correspondence, the style transfer based on such a correspondence is built. Color and contrast transfer usually changes the dominant color and contrast distribution, and maps to desirable color and contrast casts. A slightly more general approach is to fit the color statistic of the input image into that of the reference one. Global methods based on the color statistic cannot handle some tough cases, such as the image containing complex details and diverse colors. Based on the SuperBIG framework, the styles of an image could be transferred locally at the granularity of superpixel.

SuperBIG transfers colors by manipulating the statistic in the $l\alpha\beta$-CIE space, a de-correlated color space, as our local mapping method. Here we define $\left[\mathbf{I}_L,\mathbf{I}_M,\mathbf{I}_S\right]^T=\mathcal{F}\left[\mathbf{I}_R,\mathbf{I}_G,\mathbf{I}_B\right]^T$, where $\mathcal{F}$ is a predefined transformation matrix and $\mathbf{I}_R,\mathbf{I}_G,\mathbf{I}_B$ are three channels of a RGB image. Then, we convert $\left[\mathbf{I}_L,\mathbf{I}_M,\mathbf{I}_S\right]^T$ to the logarithmic space,
\begin{equation}
\mathbf{I}_\mathbf{L} = log \mathbf{I}_L, \text{ }
\mathbf{I}_\mathbf{M} = log \mathbf{I}_M, \text{ }
\mathbf{I}_\mathbf{S} = log \mathbf{I}_S, \text{ } \nonumber
\end{equation}
\begin{equation}
\left[
\begin{array}{c}
\mathbf{I}_l\\
\mathbf{I}_\alpha\\
\mathbf{I}_\beta\\
\end{array}
\right]
= \left[
\begin{array}{ccc}
\frac{1}{\sqrt{3}} & 0 & 0\\
0 & \frac{1}{\sqrt{6}} & 0\\
0 & 0 & \frac{1}{\sqrt{2}}\\
\end{array}
\right]
\left[
\begin{array}{ccc}
1 & 1 & 1\\
1 & 1 & -2\\
1 & -1 & 0\\
\end{array}
\right]
\left[
\begin{array}{c}
\mathbf{I}_\mathbf{L}\\
\mathbf{I}_\mathbf{M}\\
\mathbf{I}_\mathbf{S}\\
\end{array}
\right]. \nonumber
\end{equation}

This decorrelation makes three color channels independent. SuperBIG then adjusts the color statistic in such space by matching mean and variance as follows,
\begin{eqnarray}
\label{eq:color_transfer}
\mathbf{I}_l^{\star}&=&\mathbf{I}_l-\left<\mathbf{I}_l\right>,\text{ }\mathbf{I}_\alpha^{\star}=\mathbf{I}_\alpha-\left<\mathbf{I}_\alpha\right>,\text{ }\mathbf{I}_\beta^{\star}=\mathbf{I}_\beta-\left<\mathbf{I}_\beta\right>,\text{ } \nonumber \\
\mathbf{I}_l^{'} &=& \frac{\sigma_r^l}{\sigma_c^l}\mathbf{I}_l^{\star},\text{ }\mathbf{I}_\alpha^{'} = \frac{\sigma_r^\alpha}{\sigma_c^\alpha}\mathbf{I}_\alpha^{\star},\text{ }\mathbf{I}_\beta^{'} = \frac{\sigma_r^\beta}{\sigma_c^\beta}\mathbf{I}_\beta^{\star},
\end{eqnarray}
where $\left<\cdot\right>$ is the operator to calculate the mean and $\sigma$ is the variance of the image for a given channel. With the de-correlated style transfer for local regions, SuperBIG transfers styles between each pair of estimated corresponding superpixel pairs in $\mathbf{F}^{c}_f$ and $\mathbf{F}^{r}_f$. To avoid the boundary effect between superpixels, we finally smooth the transferred results by the guided image filter~\cite{he2010guided}.

\section{Experimental Results}
\label{sec:Exp}
\subsection{Experimental Setting}
We compare the proposed method (SuperBIG) with the following six state-of-the-art style/color transfer methods: L$\alpha\beta$ decorrelated color space (L$\alpha\beta$)~\cite{reinhard2001color}, color ``mood'' transfer (MoodTrans)~\cite{welsh2002transferring}, multi-scale harmonization (Harmonization)~\cite{sunkavalli2010multi}, landmark sparse color representation (Landmark)~\cite{huang2009landmark}, neural algorithm of artistic style (NeutralArt)~\cite{NNart} and superpixels matching (SuperMatch)~\cite{gupta2012image}. Results of these methods are generated by the published codes kindly provided by the authors. When compared to the colorization methods, SuperBIG first turns the input image into greyscale one, then colorizes the generated greyscale image. We set the parameters as: $\lambda_{\mathbf{M}}=0.1, \lambda_{\mathbf{T}}=0.001, \lambda_{\mathbf{C}}=0.0001, \lambda_{\mathbf{DV}}=10^{-6}, \lambda_{\mathbf{S}}=0.1, \lambda_{\mathbf{L^a}}=\lambda_{\mathbf{L^r}}=0.01, n_{\pmb{\alpha}}=1000$ and $n_{\pmb{\beta}}=10^6$.

\subsection{Comparison with State-of-the-Art \\and Various Styles}
The comparison results of SuperBIG and other state-of-the-art methods for three input images are presented in Figures~\ref{fig:Lourvre2-1}-\ref{fig:Taj3}. Please enlarge and view these figures on the screen for better comparison. The subjective quality of these results demonstrates the superiority of the proposed SuperBIG. L$\alpha\beta$ and Harmonization totally fail to transfer the color, because of wrong dominant color prediction in Figures~\ref{fig:Lourvre2-1}(b) and \ref{fig:Taj1}(b) as well as heavily blurred or extremely rough sky regions in Figures~\ref{fig:Lourvre2-1}(c)-\ref{fig:Taj3}(c), respectively. Landmark, NeutralArt and SuperMatch suffer from wrong local style predictions, \emph{e.g.} blue color near the edges and corners of the pyramid in Figures~\ref{fig:Lourvre2-1}(d)(f)(g) and the color artifacts on the top of the towers of Taj Mahal in Figures~\ref{fig:Taj3}(d)(f)(g).
Thanks to informative hierarchical features and effective superpixel bipartite framework for modeling in the global and local correspondences, SuperBIG transfers the proper styles for the local regions in the generated results as shown in Figures~\ref{fig:Lourvre2-1}(h)-\ref{fig:Taj3}(h).

\def\height{2.7cm}
\begin{figure*}[htbp]
	\begin{center}
		{
			\subfigure[Input]{
				\includegraphics[height=\height,clip]{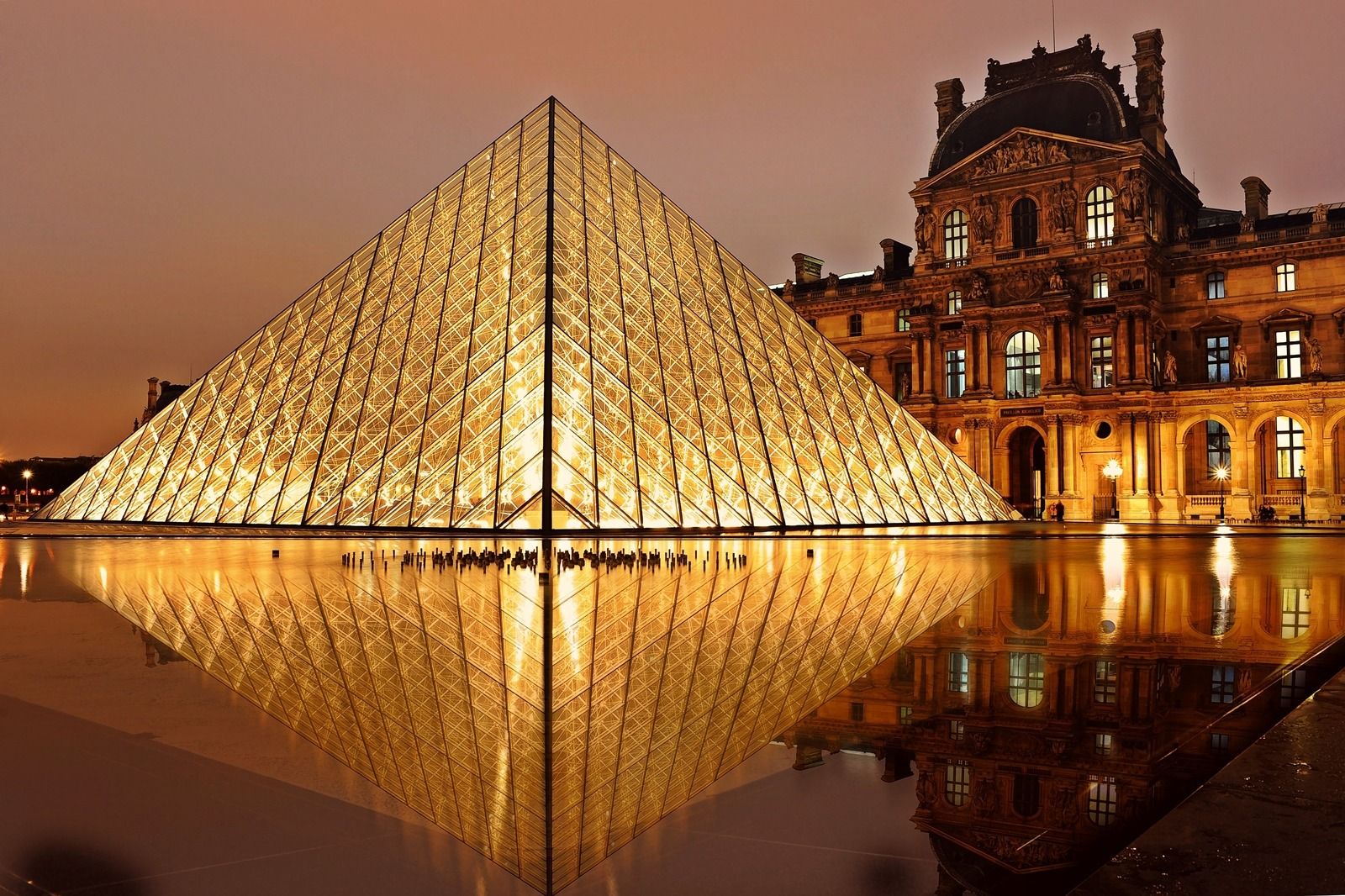}
			}
			\subfigure[L$\alpha\beta$]{
				\includegraphics[height=\height,clip]{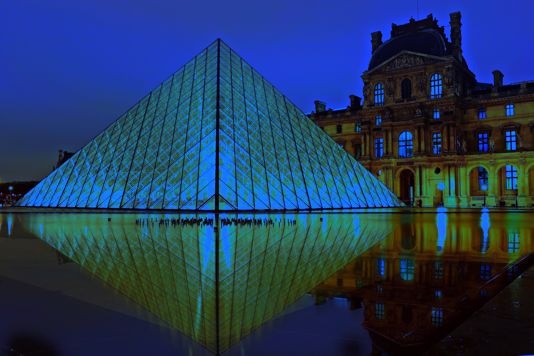}
			}
			\subfigure[Harmonization]{
				\includegraphics[height=\height,clip]{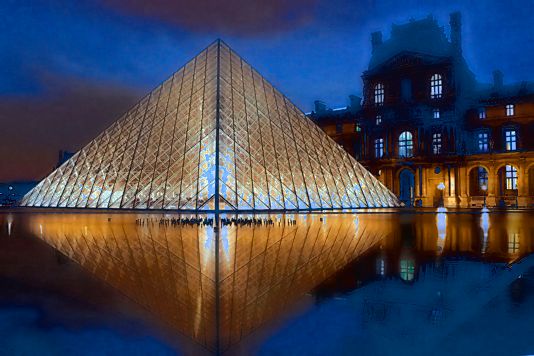}
			}
			\subfigure[Landmark]{
				\includegraphics[height=\height,clip]{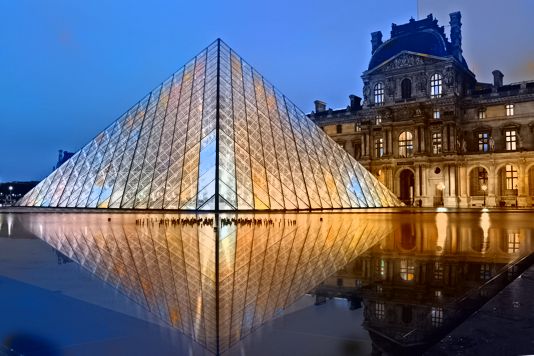}
			}
			\\
			\subfigure[Reference]{
				\includegraphics[height=\height,clip]{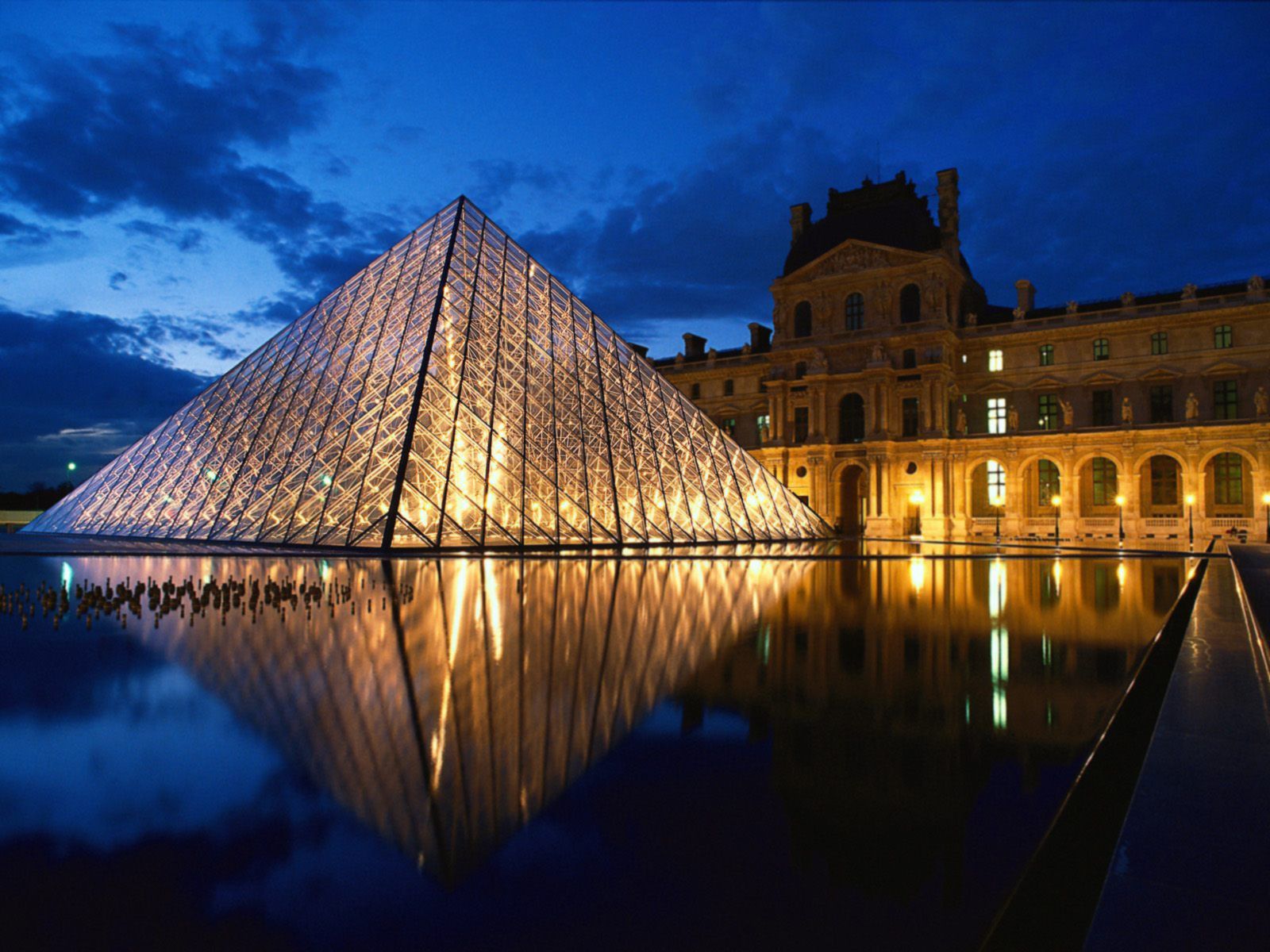}
			}			
			\subfigure[NeutralArt]{
				\includegraphics[height=\height,clip]{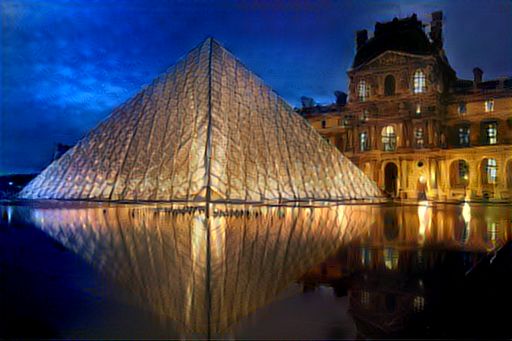}
			}
			\subfigure[SuperMatch]{
				\includegraphics[height=\height,clip]{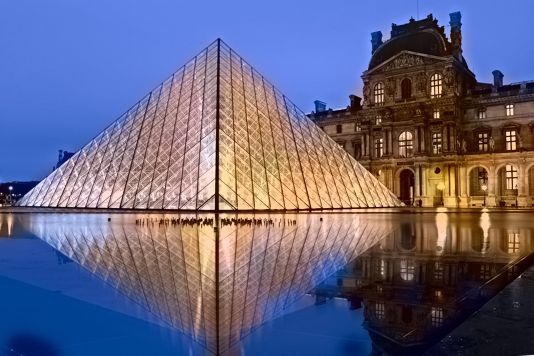}
			}
			\subfigure[SuperBig]{
				\includegraphics[height=\height,clip]{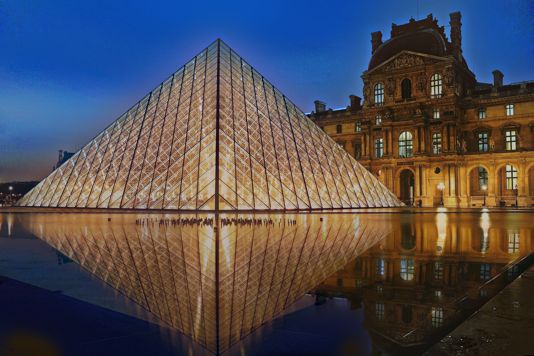}
			}
		}
	\end{center}
	\vspace{-2mm}
	\caption{ Visual comparisons of style transfer from (a) to (e) among different algorithms.}
	\label{fig:Lourvre2-1}
\end{figure*}

\def\hs{-0.3cm}
\def\height{2.7cm}
\begin{figure*}[htbp]
	\begin{center}
		{
			\subfigure[Input]{
				\includegraphics[height=\height,clip]{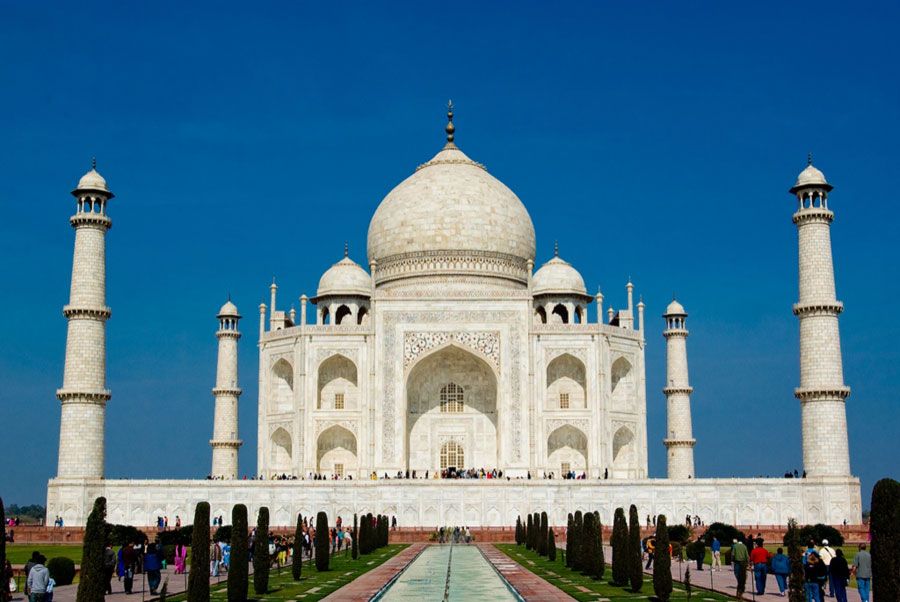}
			}
			\hspace{\hs}
			\subfigure[L$\alpha\beta$]{
				\includegraphics[height=\height,clip]{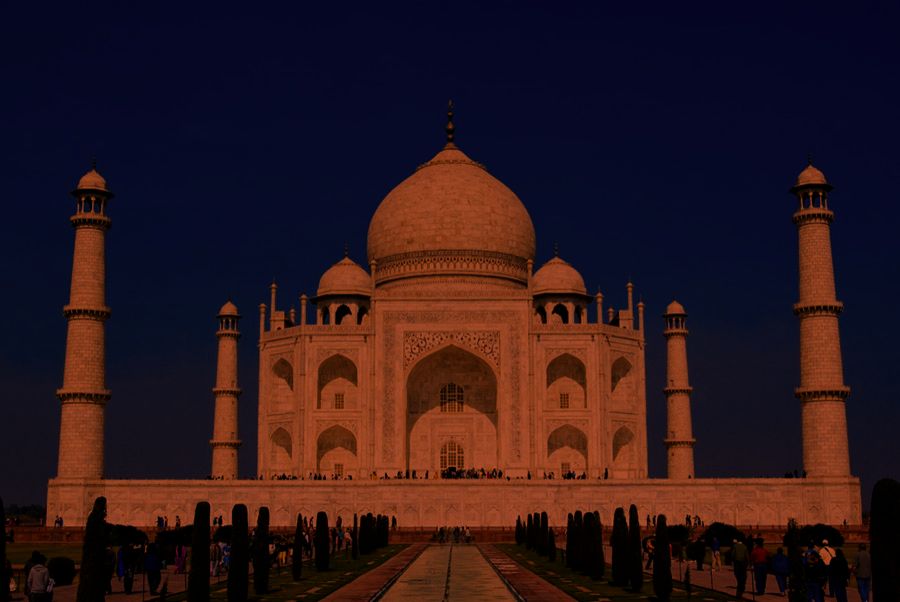}
			}
			\hspace{\hs}
			\subfigure[Harmonization]{
				\includegraphics[height=\height,clip]{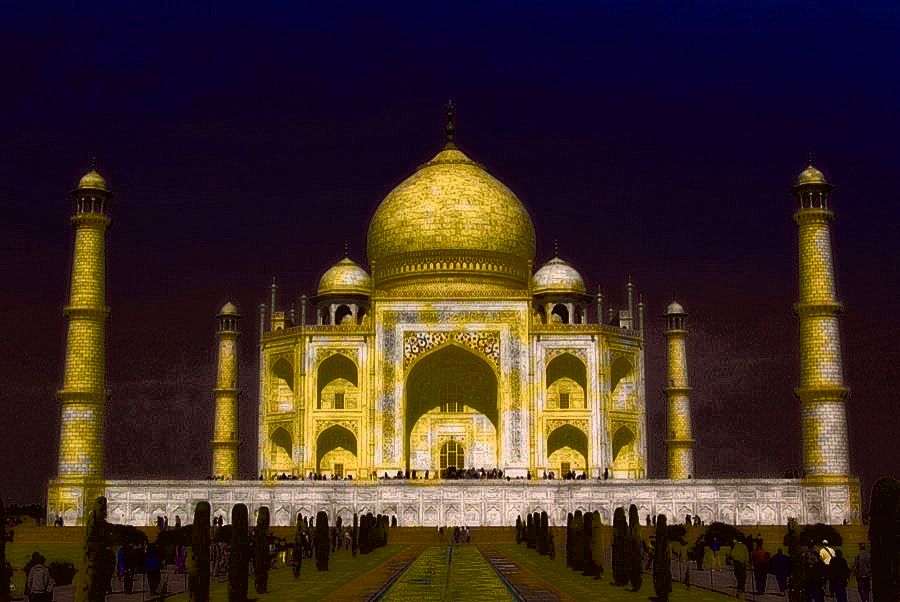}
			}
			\hspace{\hs}
			\subfigure[Landmark]{
				\includegraphics[height=\height,clip]{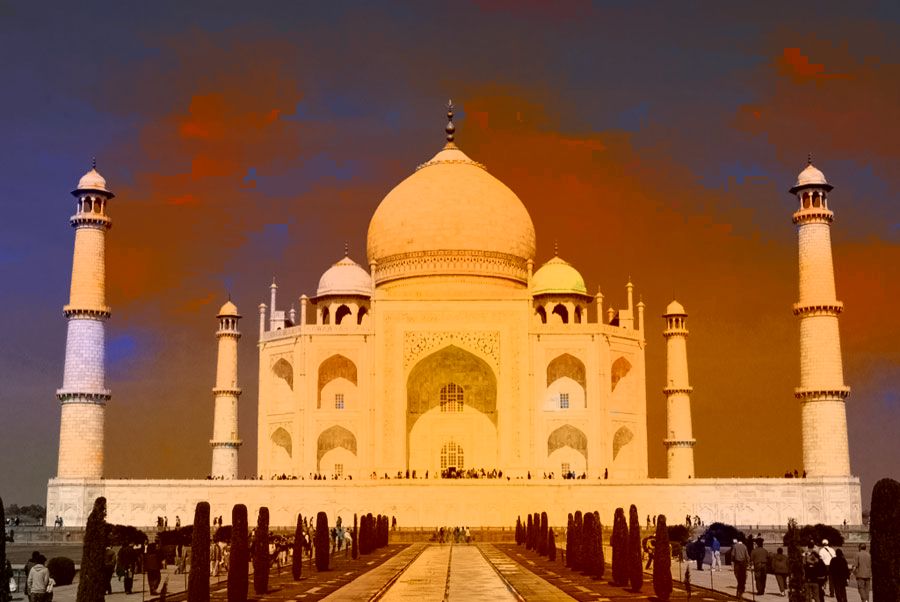}
			}
			\\
			\vspace{-3mm}
			
			\subfigure[Reference]{
				\includegraphics[height=\height,clip]{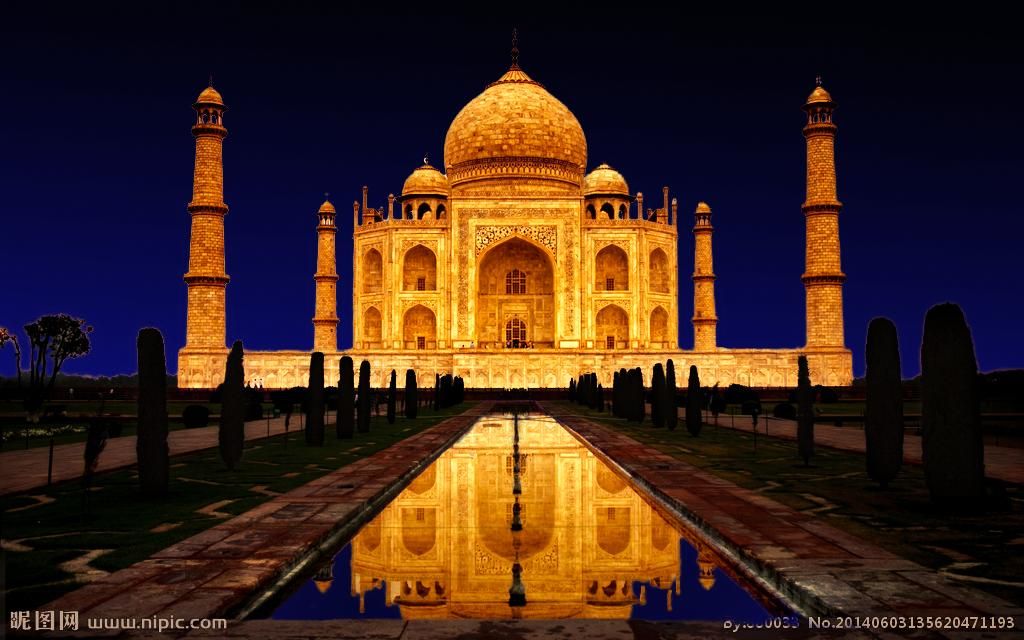}
			}
			\hspace{\hs}		
			\subfigure[NeutralArt]{
				\includegraphics[height=\height,clip]{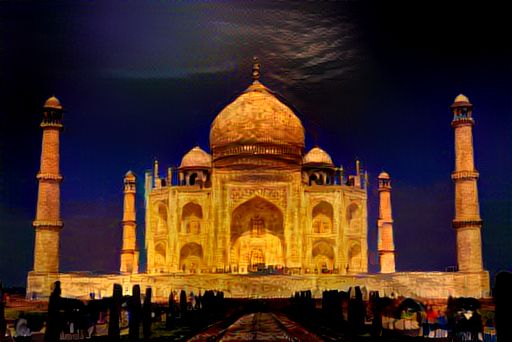}
			}
			\hspace{\hs}
			\subfigure[SuperMatch]{
				\includegraphics[height=\height,clip]{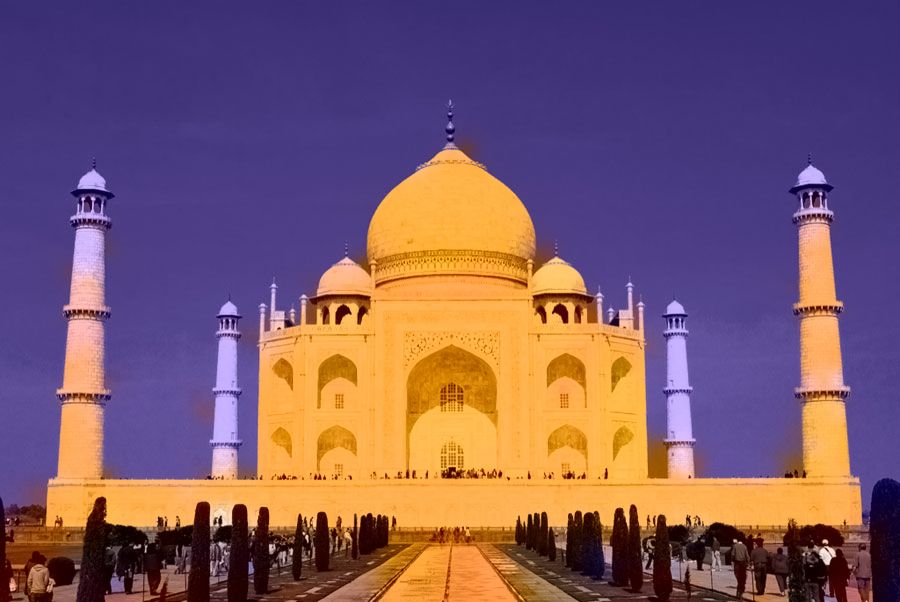}
			}
			\hspace{\hs}
			\subfigure[SuperBig]{
				\includegraphics[height=\height,clip]{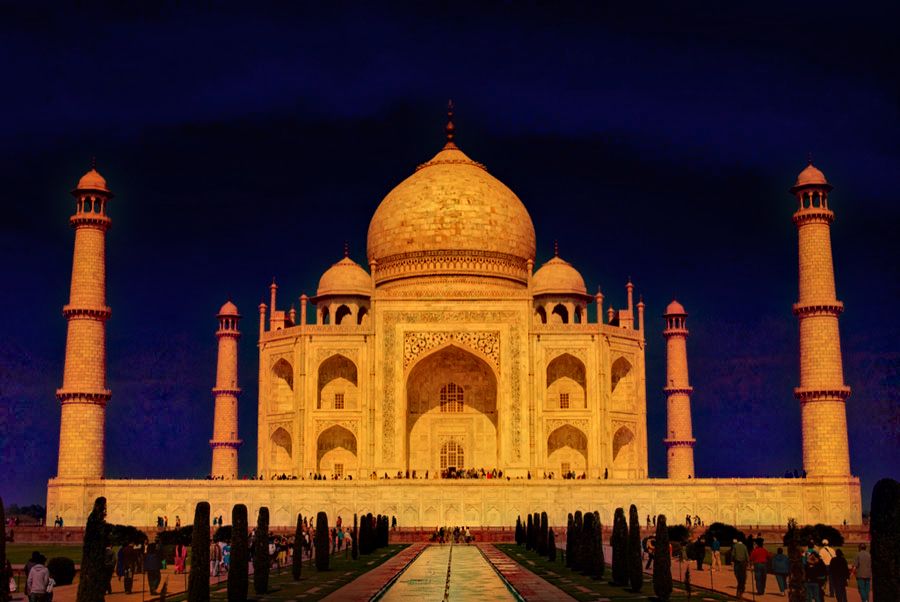}
			}
		}
	\end{center}
	\vspace{-3mm}
	
	\caption{Visual comparisons of style transfer from (a) to (e) among different algorithms.}
	\label{fig:Taj1}
\end{figure*}
\def\hs{-0.3cm}
\def\height{2cm}
\begin{figure*}[htbp]
	\begin{flushright}
		{
			\subfigure[Input]{
				\includegraphics[height=\height,clip]{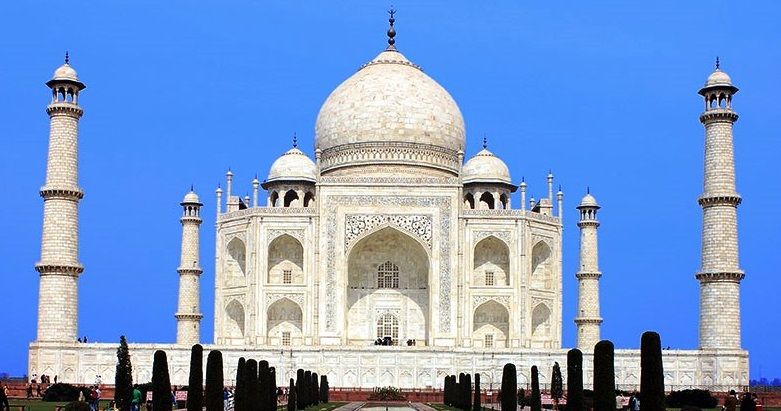}
			}
			\hspace{\hs}				
			\subfigure[L$\alpha\beta$]{
				\includegraphics[height=\height,clip]{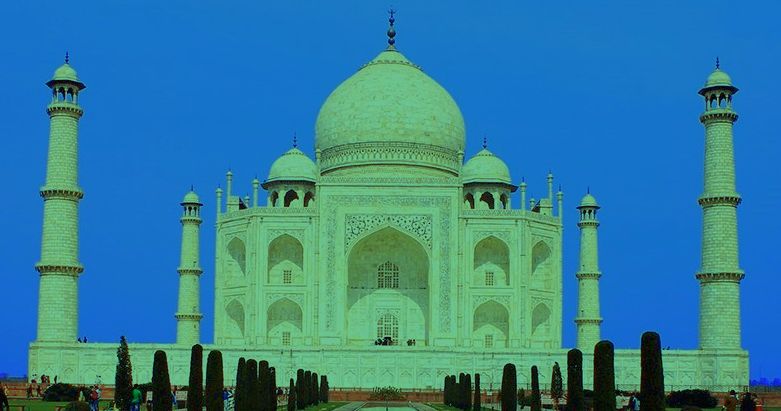}
			}
			\hspace{\hs}			
			\subfigure[Harmonization]{
				\includegraphics[height=\height,clip]{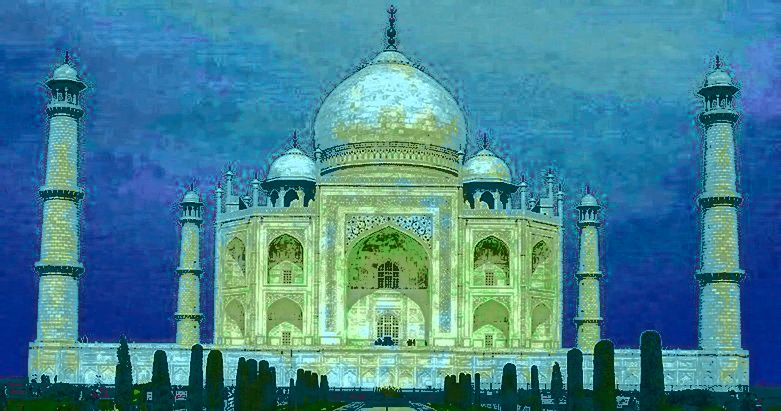}
			}
			\hspace{\hs}			
			\subfigure[Landmark]{
				\includegraphics[height=\height,clip]{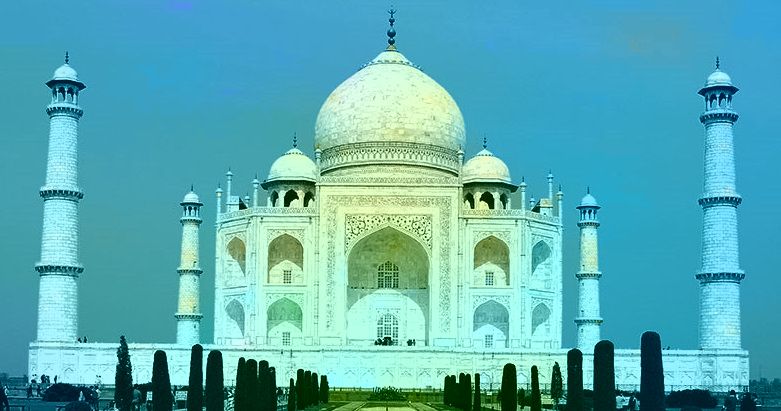}
			}
			\\
			\vspace{-3mm}
			
			\subfigure[Reference]{
				\includegraphics[height=\height,clip]{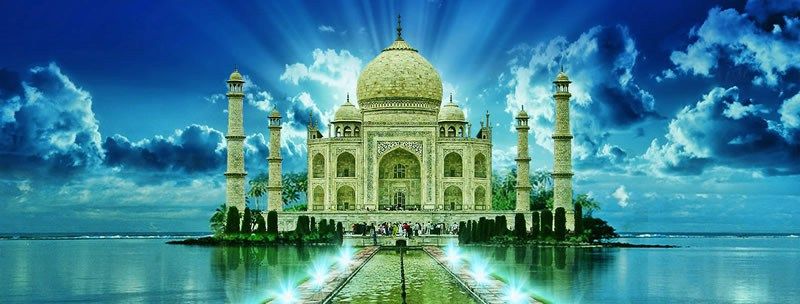}
			}
			\hspace{\hs}					
			\subfigure[NeutralArt]{
				\includegraphics[height=\height,clip]{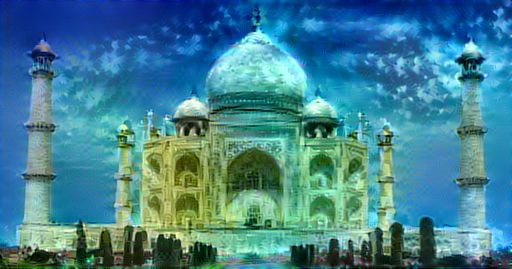}
			}
			\hspace{\hs}			
			\subfigure[SuperMatch]{
				\includegraphics[height=\height,clip]{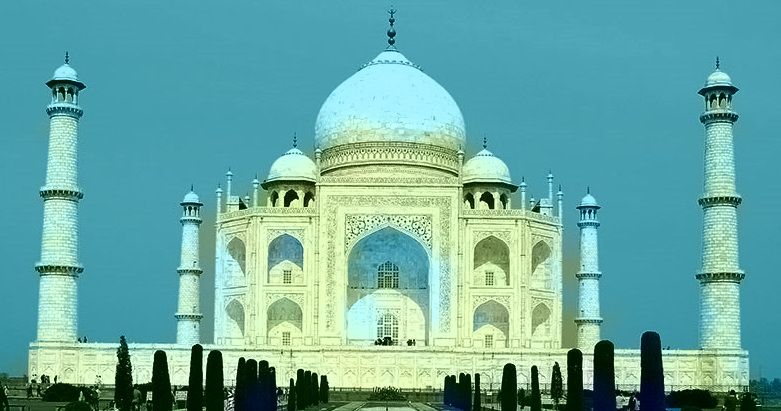}
			}
			\hspace{\hs}			
			\subfigure[SuperBig]{
				\includegraphics[height=\height,clip]{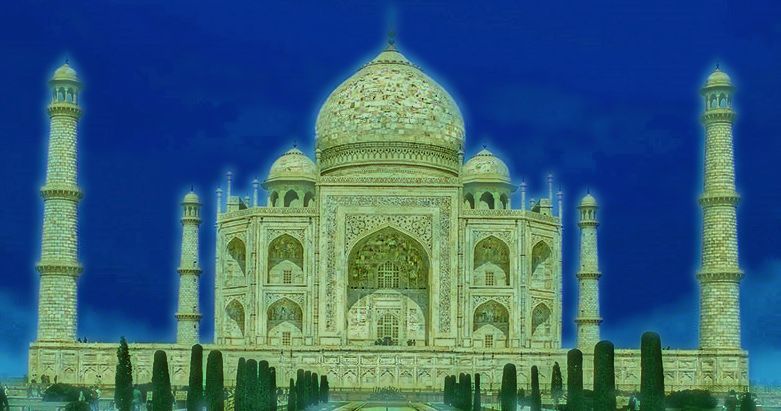}
			}
		}
	\end{flushright}
	\caption{ Visual comparisons of style transfer from (a) to (e) among different algorithms.}
	\vspace{-3mm}
	
	\label{fig:Taj3}
\end{figure*}

The subjective results of SuperBIG to transfer different styles are showed in Figure~\ref{fig:all1}. From the results, we observe that SuperBIG generates the results containing clear and natural content while successfully changing their styles based on the reference images, leading to similar spatial distribution of color and contrast. It is worth noting that, even for the night image as shown in the right-bottom of Figure~\ref{fig:all1}(b), where background light is dim, SuperBIG can still achieve the transformation successfully and generate naturally looking results.

\def\heighto{34mm}
\def\heightot{34mm}
\def\heightt{9mm}
\def\heighttt{6mm}
\def\hstart{0}
\def\hs{-1mm}

\begin{figure*}[htbp]
	\begin{center}
		{
			\includegraphics[width=\heighto,clip]{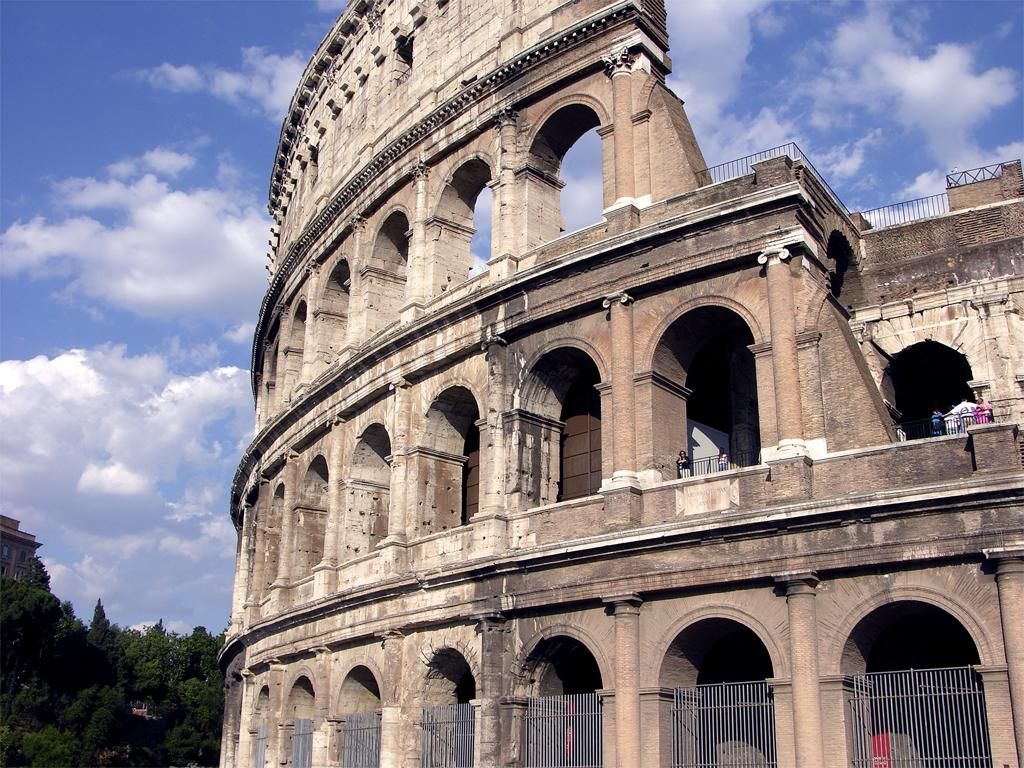}
			\hspace{\hs} 			
            	\begin{overpic}[width=\heighto,clip]{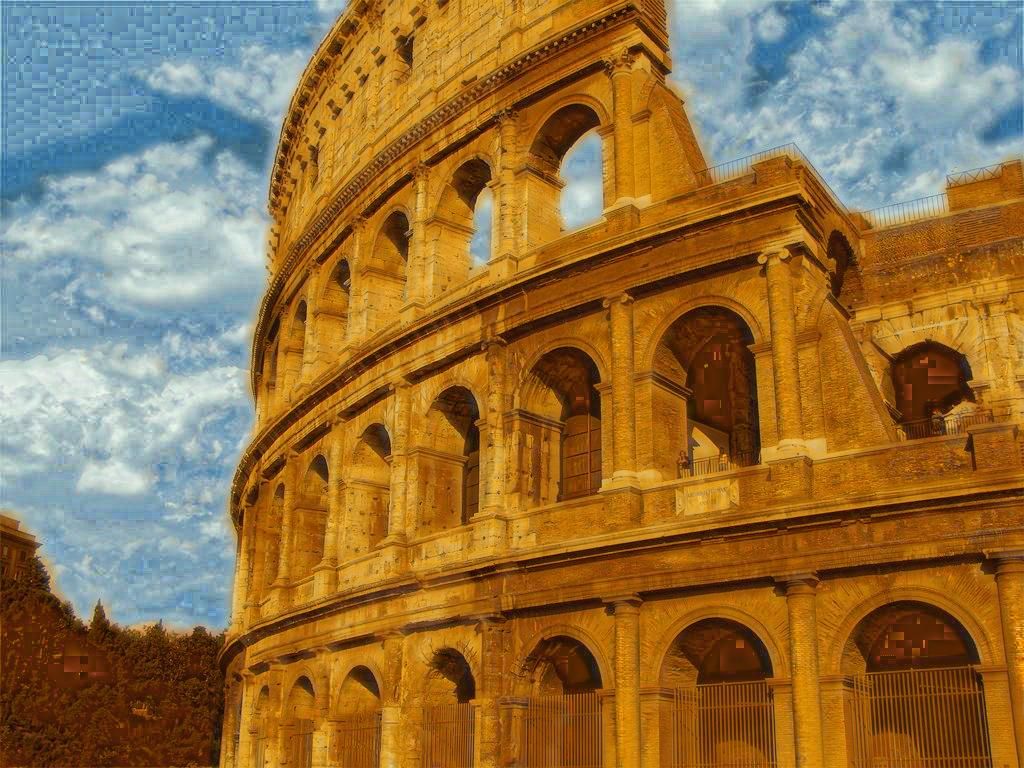}
            		\put(0,\hstart){\includegraphics[height=\heightt,clip]{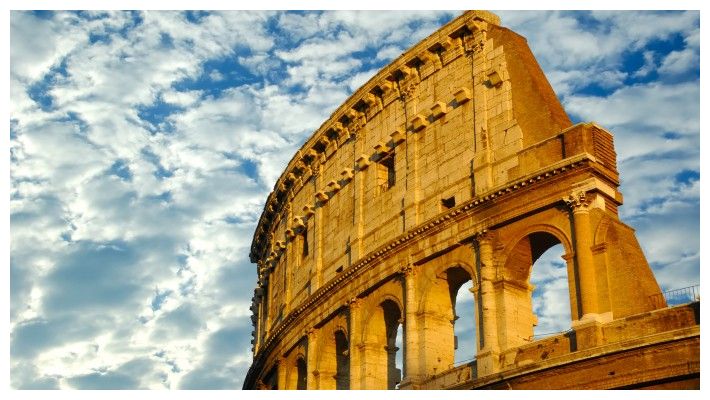}}
            	\end{overpic}
	            \hspace{\hs}
            	\begin{overpic}[width=\heighto,clip]{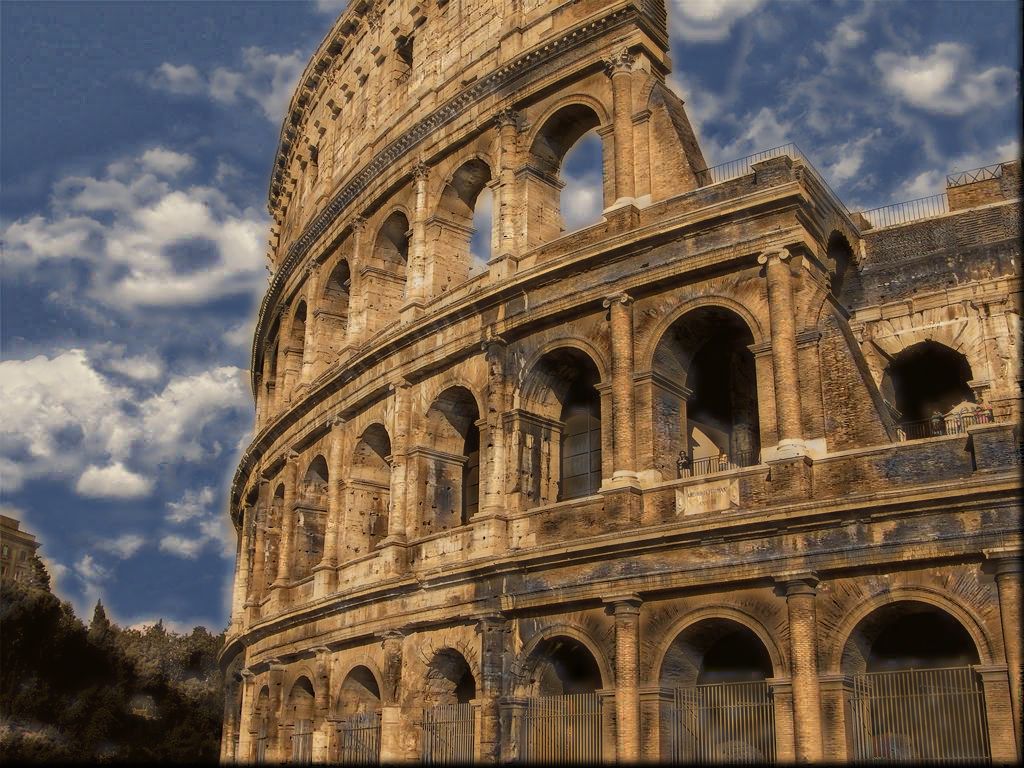}
            		\put(0,\hstart){\includegraphics[height=\heightt,clip]{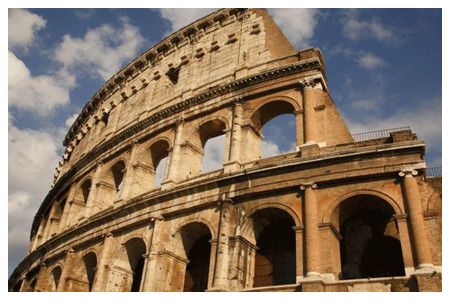}}
            	\end{overpic}
	            \hspace{\hs}
            	\begin{overpic}[width=\heighto,clip]{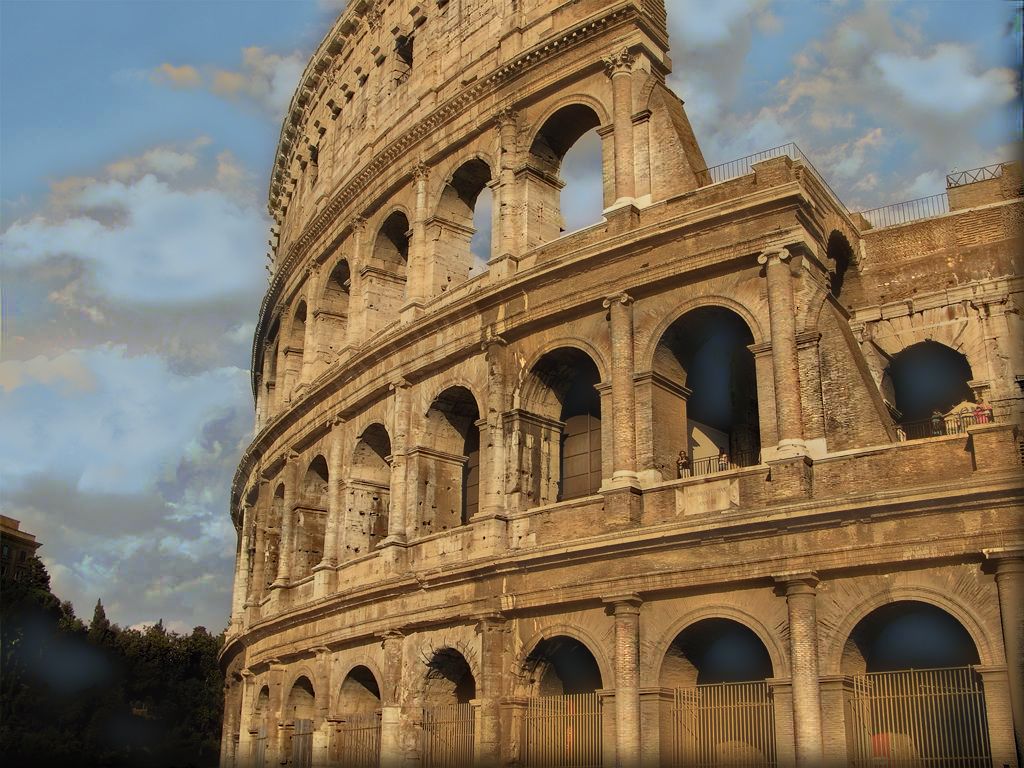}
            		\put(0,\hstart){\includegraphics[height=\heightt,clip]{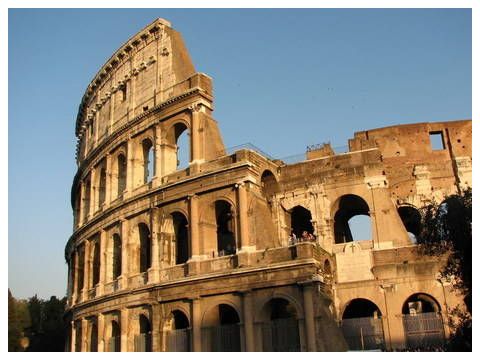}}
            	\end{overpic}
	            \hspace{\hs}
            	\begin{overpic}[width=\heighto,clip]{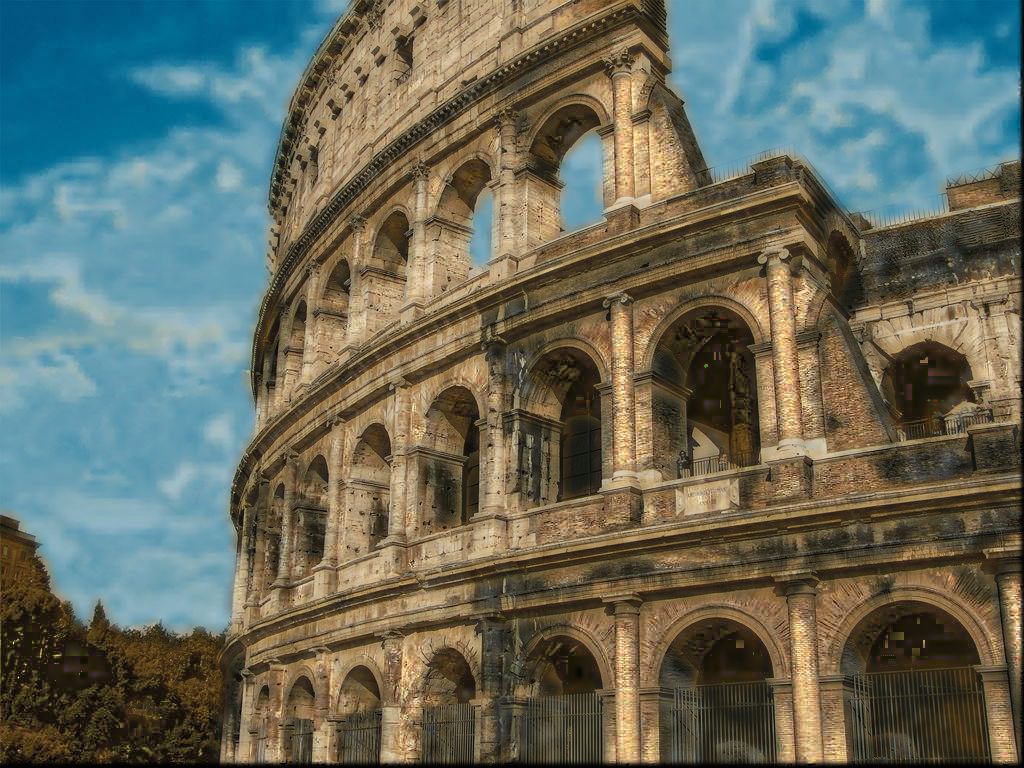}
            		\put(0,\hstart){\includegraphics[height=\heightt,clip]{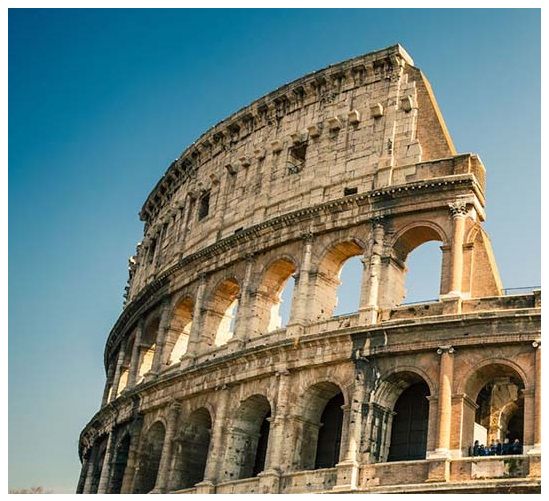}}
            	\end{overpic}
            \\
            \vspace{-1mm}
			\subfigure[\textbf{Input}]{
				\includegraphics[width=\heightot,clip]{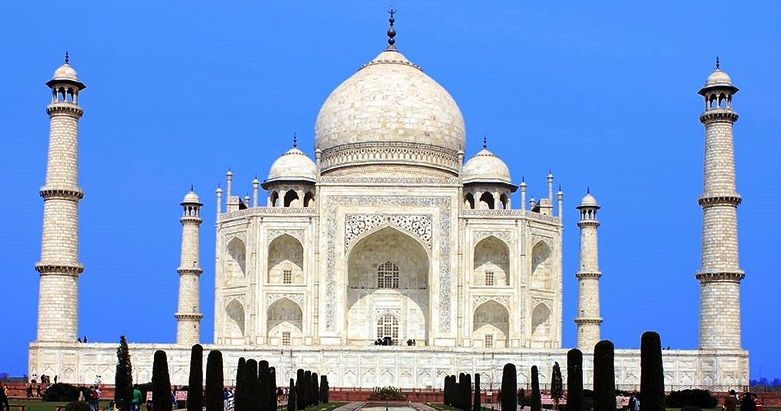}
			}
			\hspace{-3mm}
			\subfigure[\textbf{Output}: Styles transferred photos from the examples. The inserts show the examples.]{
				\begin{overpic}[width=\heightot,clip]{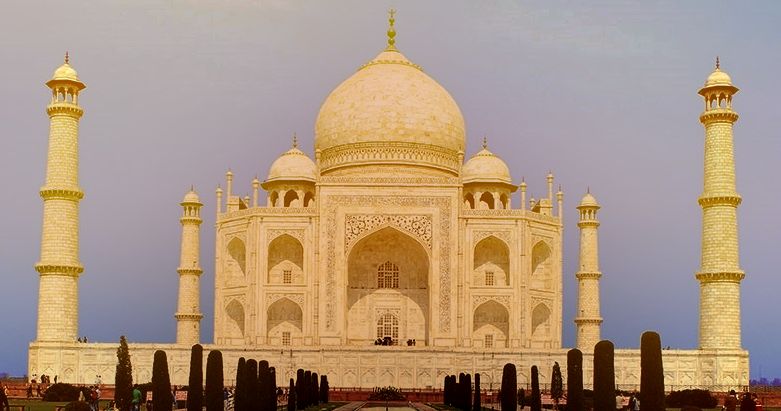}
					\put(0,\hstart){\includegraphics[height=\heighttt,clip]{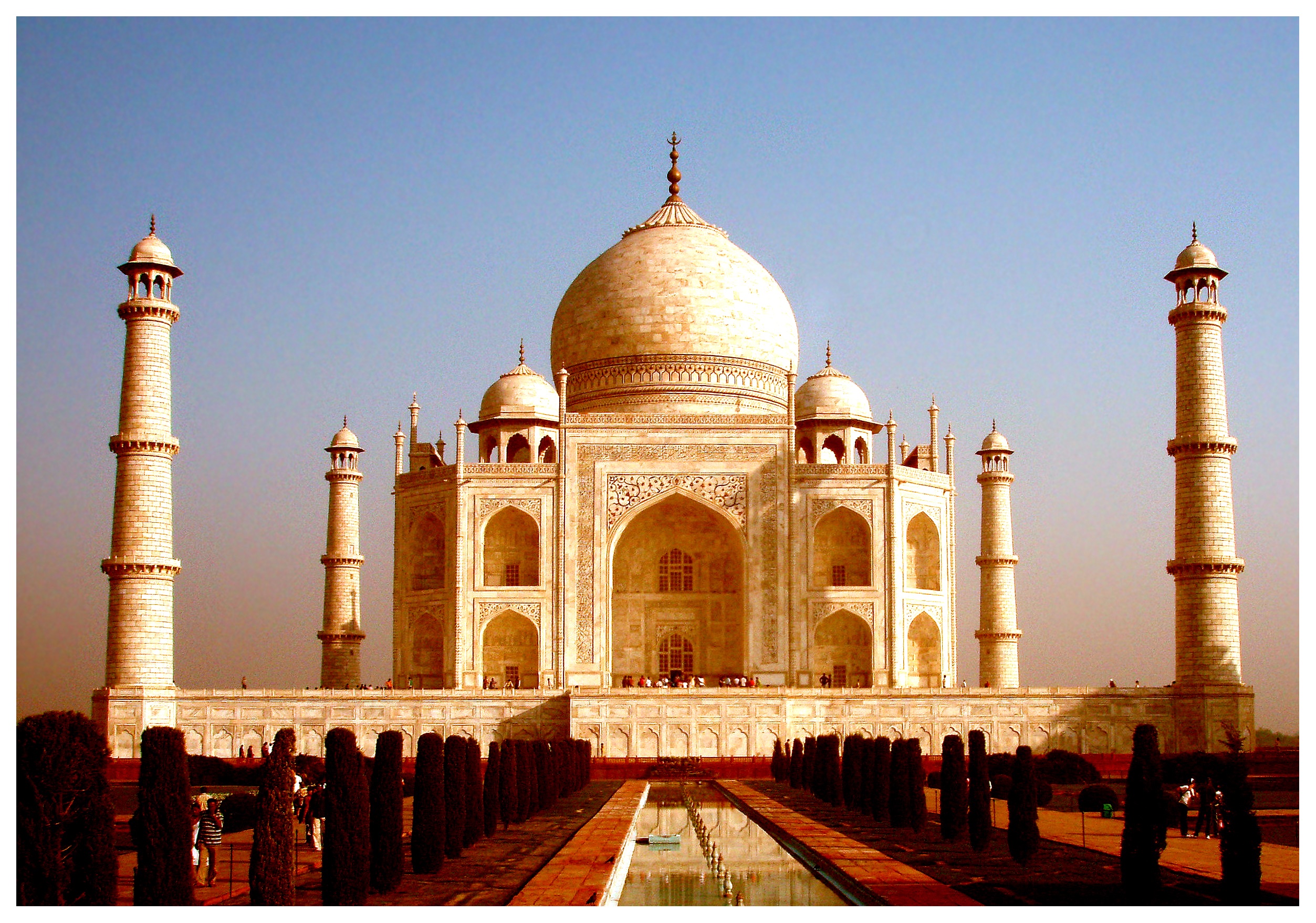}}
				\end{overpic}
			\hspace{\hs}
				\begin{overpic}[width=\heightot,clip]{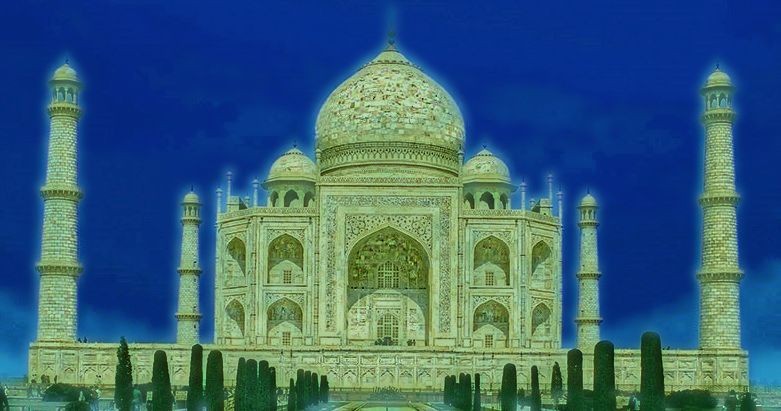}
					\put(0,\hstart){\includegraphics[height=\heighttt,clip]{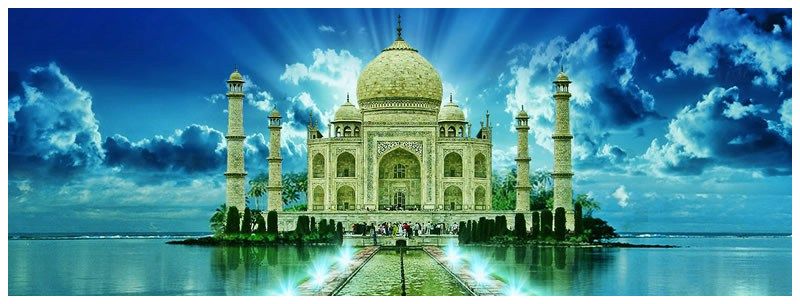}}
				\end{overpic}
			\hspace{\hs}
				\begin{overpic}[width=\heightot,clip]{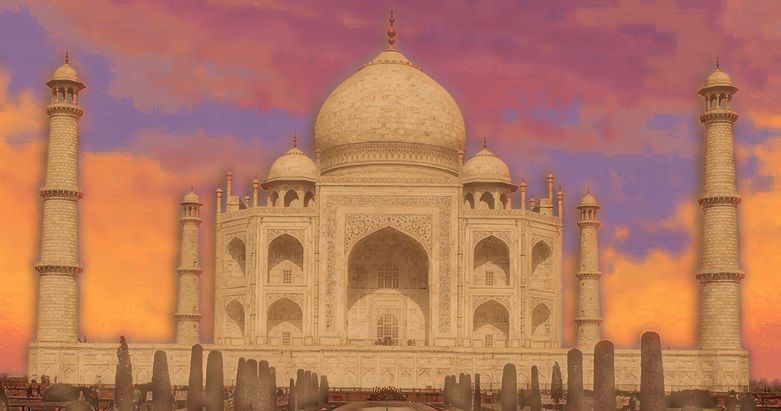}
					\put(0,\hstart){\includegraphics[height=\heighttt,clip]{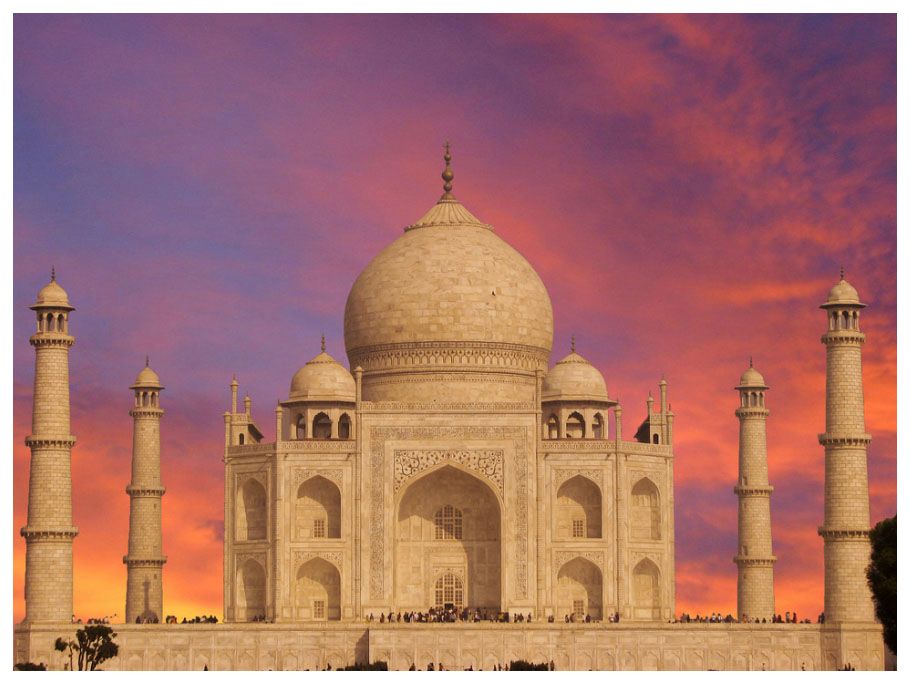}}
				\end{overpic}
			\hspace{\hs}
				\begin{overpic}[width=\heightot,clip]{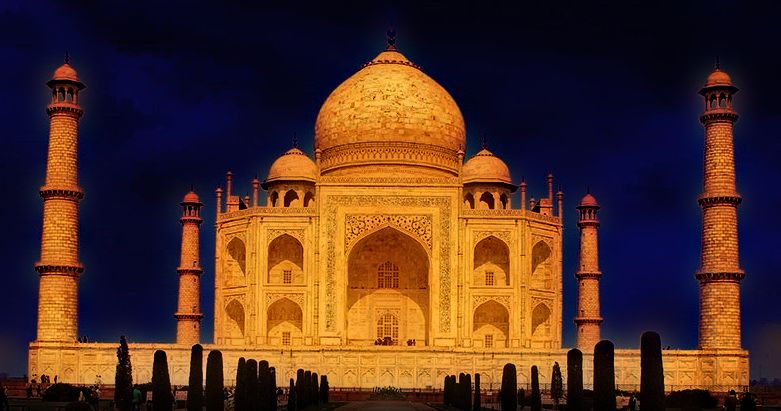}
					\put(0,\hstart){\includegraphics[height=\heighttt,clip]{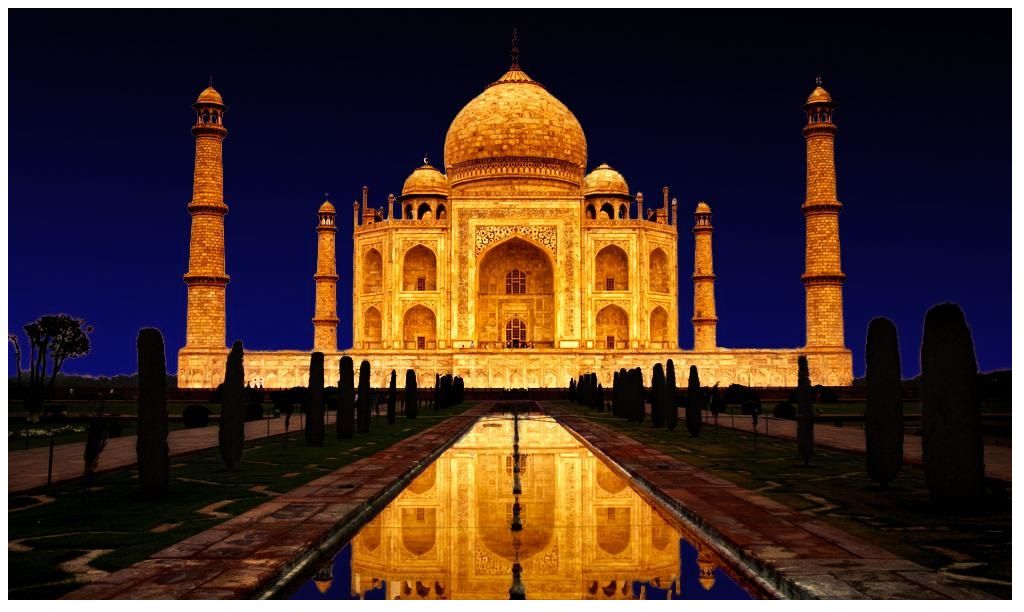}}
				\end{overpic}
			}
			                       			
%
%
		}
	\end{center}
	\vspace{-5mm}
	\caption{ Visual comparisons of SuperBIG style transfer for different reference images.}
	\label{fig:all1}
\end{figure*}

\subsection{User Study in Subjective Evaluation}
\begin{figure*}[htb]
	\centering\includegraphics[scale=0.6]{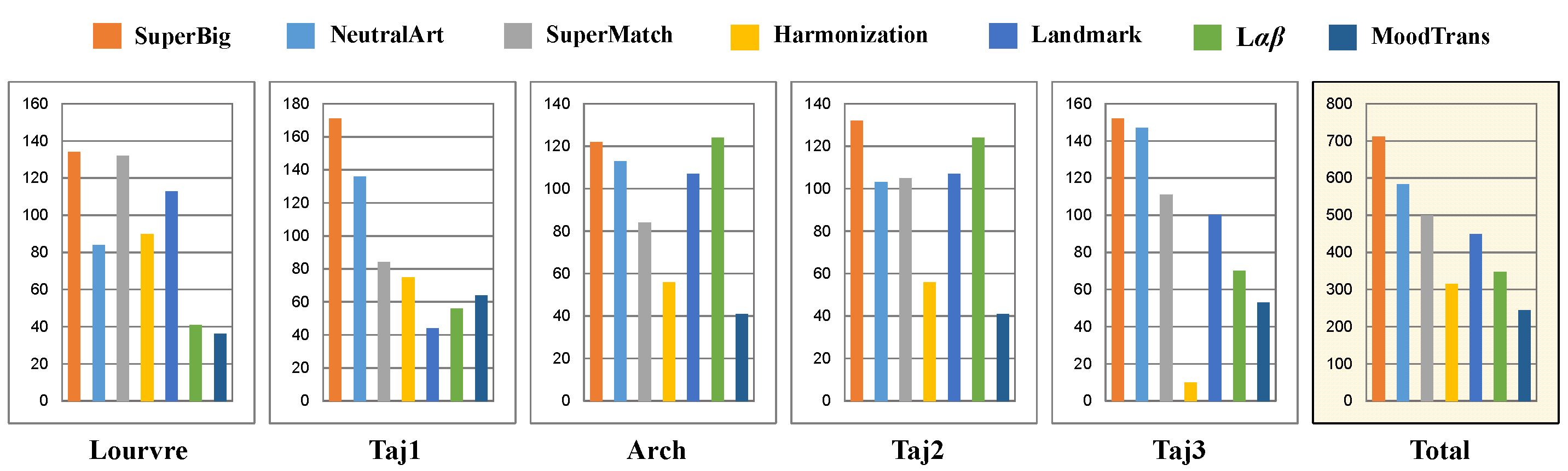}
    \vspace{-3mm}
	\caption{The number of votes per testing image and the total ranking of seven methods.}
	\label{fig:subjective}
\end{figure*}

To compare different stylization results from an observer's perspective, we employ the paired comparisons approach, where the participants are shown two stylized images at a time, side by side, and are asked to simply choose the preferred one by considering both visual quality and similar style to the exemplar. We have a total of 90 participants, including both domain experts and generally knowledgeable individuals, each given 105 pairwise comparisons over a set of five images with seven different style transfer methods. Figure~\ref{fig:subjective} illustrates the seven methods, ranked by the number of votes received. It can be seen that the proposed SuperBIG outperforms other methods in four out of the five cases, and achieves overall superior performance. Even in the exceptional case with the test image \textit{Arch}, it still shows comparable performance with the first ranked method. Besides the voting statistic, we also show the stability analysis, which is calculated by the \emph{rank product}~\cite{rubinstein2010comparative}. Table~\ref{tab:rank} shows the results of the rank product $\psi(O)=\left(\prod_i r_{O,i}\right)^{1/b}$, where $r_{O,i}$ is the specific ranking for method $O$ and image $i~(i=1\ldots b)$. Compared with others, SuperBIG produces the best consistency among different test cases to achieve the best visual quality.
\begin{table*}[htb]
\caption{Comparison of the rank product of seven methods.} \vspace{1mm}
\label{tab:rank}
\centering
\begin{tabular} {|c|c|c|c|c|c|c|c|c|}\hline
Method        & CIE-L$\alpha\beta$ & Harmonization & Landmark & MoodTrans & NeutralArt & SuperMatch & SuperBIG          \\ \hline
Rank $\psi$   & 4.04               & 5.07          & 4.22     & 6.35      & 2.83       & 2.83       & \textbf{1.15}     \\ \hline
\end{tabular}
\end{table*}

\subsection{Ablation Analysis}
To further explore the functionality of each step of SuperBIG, we perform the ablation analysis of each step in the flowchart as shown in Figure~\ref{fig:stepall}. We find that deep matching provides a large amount of matched points. It can be observed from Figures~\ref{fig:stepall}(b)(g) that most of them are visually correct. Taking a given portion of matched points ($70\%$ with highest confidence scores) and calculating the hierarchical features, SuperBIG obtains superpixels around matched points as shown in Figures~\ref{fig:stepall}(c)(h). Afterwards, uncovered pixels are handled in a pixel-level bipartite graph to generate other superpixels in Figures~\ref{fig:stepall}(d)(i). According to the correspondence obtained so far, we generate the style transfer result of Figure~\ref{fig:stepall}(e). It can be seen that, because the matching from the previous steps does not consider the global information, it generates only the locally consistent result. There are some visually unpleasant details. First, there are some inaccurate color transfer results in the right- bottom part of the image. Second, the sky in Figure~\ref{fig:stepall}(e) presents abundant textures, different from that in both the input and reference images. Thus, SuperBIG reconsiders the matching between all superpixels of the two images. Due to the feature refined from pixels to superpixels and global optimization, SuperBIG generates a well-constructed result in Figure~\ref{fig:stepall}(j).
\def\wsp{-0.3cm}
\def\width{3.4cm}
\begin{figure*}[htbp]
	\begin{center}
		{
			\subfigure[]{
				\includegraphics[width=\width,clip]{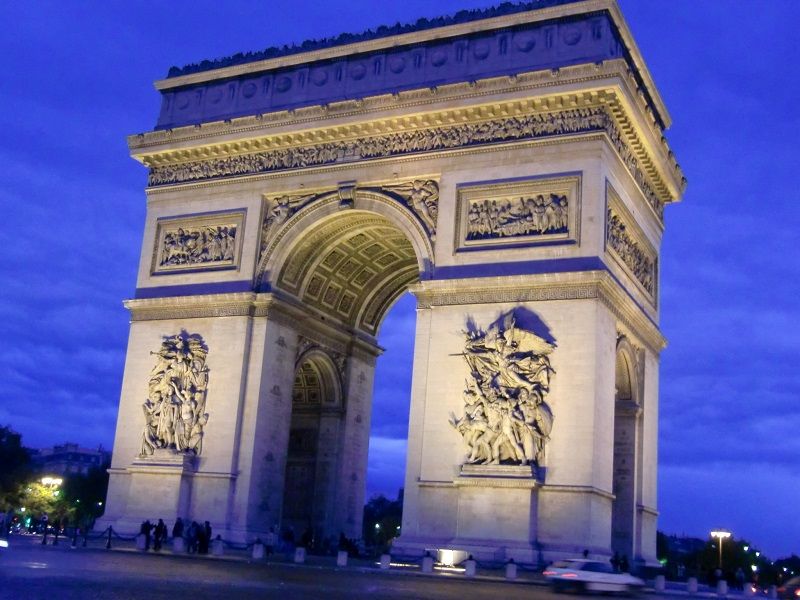}
			}
			\hspace{\wsp}
			\subfigure[]{
				\includegraphics[width=\width,clip]{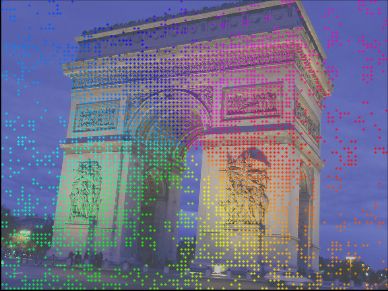}
			}
			\hspace{\wsp}
			\subfigure[]{
				\includegraphics[width=\width,clip]{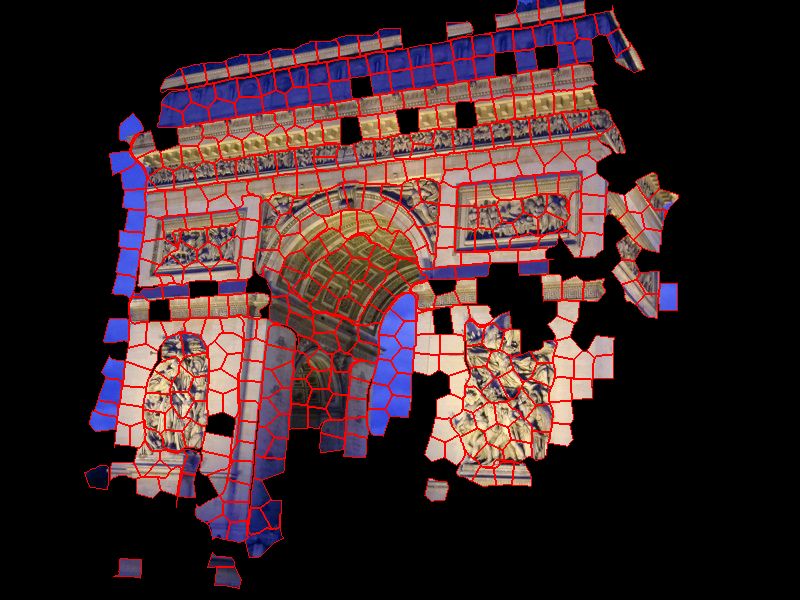}
			}
			\hspace{\wsp}
			\subfigure[]{
				\includegraphics[width=\width,clip]{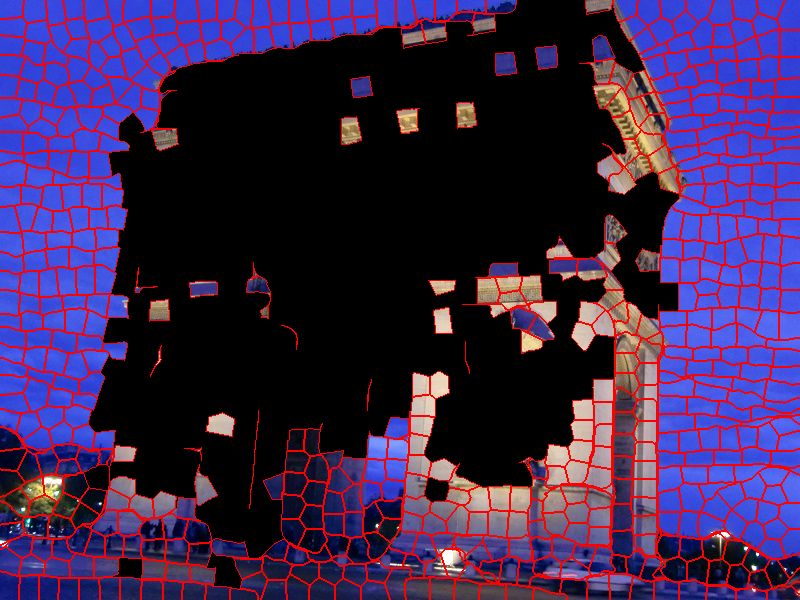}
			}	
			\hspace{\wsp}
			\subfigure[]{
				\includegraphics[width=\width,clip]{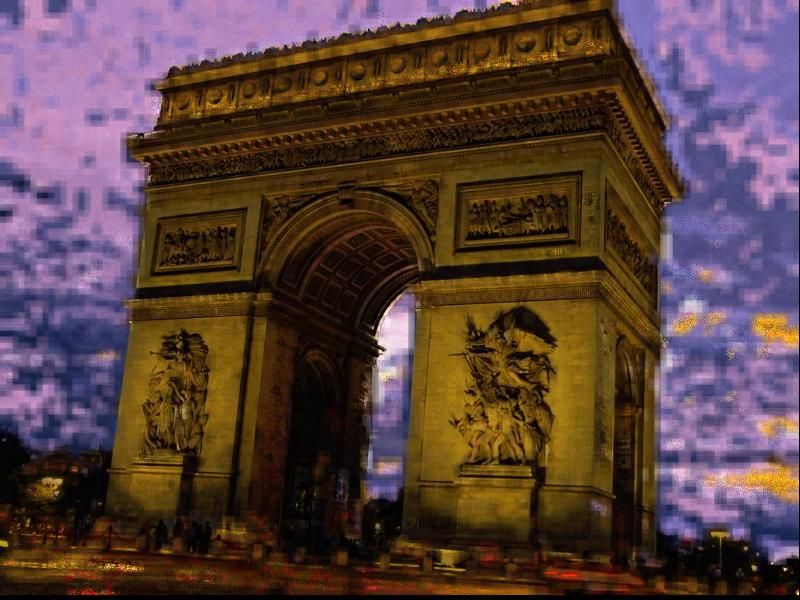}
			}						
			\\
			\vspace{-3mm}
			
			\subfigure[]{
				\includegraphics[width=\width,clip]{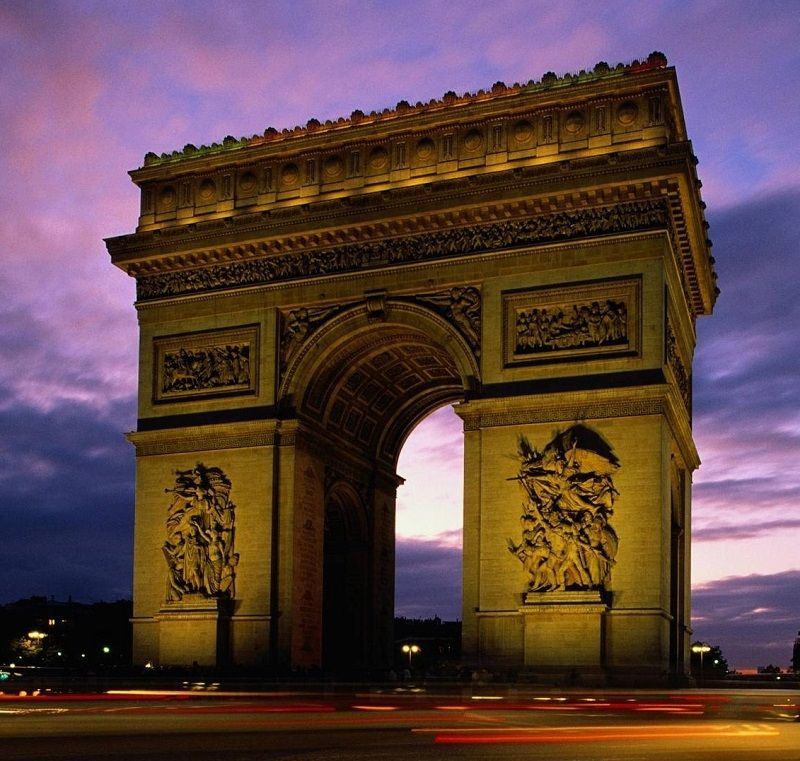}
			}
			\hspace{\wsp}
			\subfigure[]{
				\includegraphics[width=\width,clip]{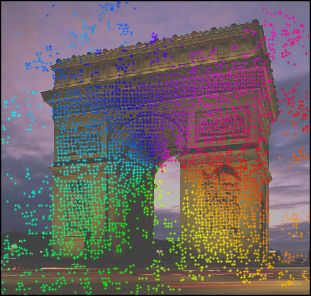}
			}			
			\hspace{\wsp}
			\subfigure[]{
				\includegraphics[width=\width,clip]{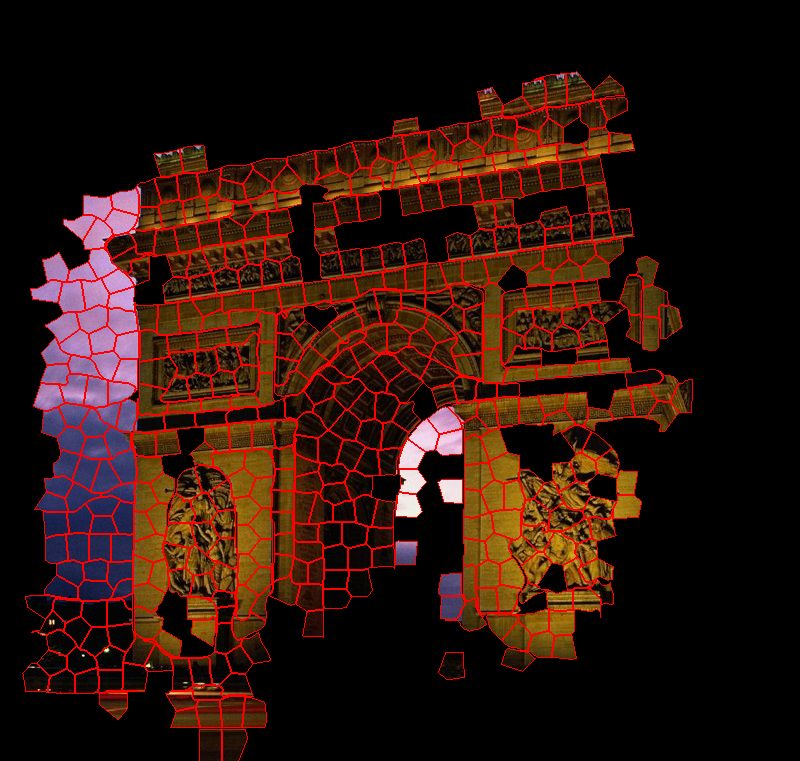}
			}
			\hspace{\wsp}
			\subfigure[]{
				\includegraphics[width=\width,clip]{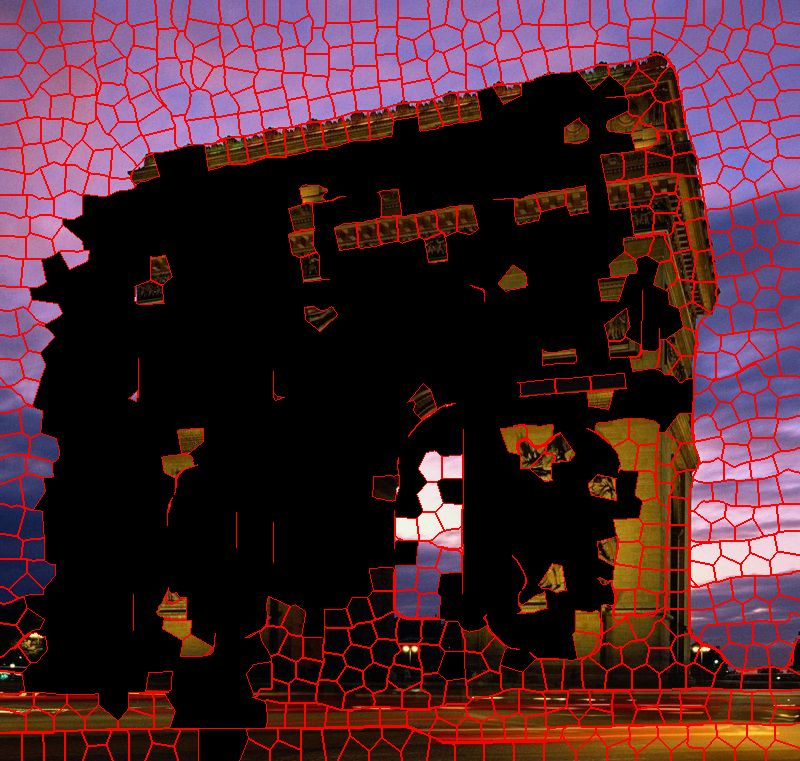}
			}
			\hspace{\wsp}				
			\subfigure[]{
				\includegraphics[width=\width,clip]{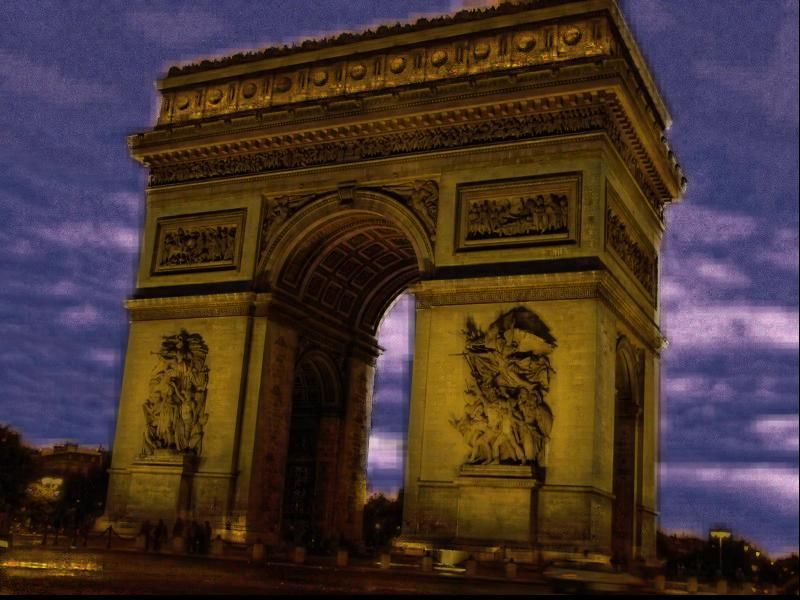}
			}																						
		}
	\end{center}
	\vspace{-6mm}
	\caption{The ablation analysis for SuperBIG. (a) The input image. (b) Dense correspondence in (a). (c) Superpixels for matched points in (a). (c) Superpixels for other pixels in (a). (e) The transfer results with the superpixel correspondece generated from the pixel-level bipartite graph partition. (f) The reference image. (g) Dense correspondence in (f). (h) Superpixels for matched points in (f). (i) Superpixels for other pixels in (f). (j) The transfer results with the superpixel correspondece generated from the superpixel-level bipartite graph matching.}
	\label{fig:stepall}
\end{figure*}

\subsection{Hierarchical Features Analysis}
We also explore the effectiveness of each feature in the hierarchical features. We only focus on the functionality of primitive features: color, distance (absolute and relative), texture, patch intensity vector, gradient. Figure \ref{fig:featureall} shows the results generated by SuperBIG with the compositions of these five features. From the results, it could be seen that the composition of color and distance, or patch intensity vector alone leads to the result containing many falsely transferred regions. Adding the texture feature removes many false regions by texture consistency. However, the quality of the sky is limited. The patch intensity vector puts the local constraint on the transfer and generates naturally looking result. The gradient feature generates a more smooth result with a higher visual quality.
\def\wsp{-0.3cm}
\def\width{3.4cm}
\begin{figure*}[htbp]
	\begin{center}
		{
			\subfigure[]{
				\includegraphics[width=\width,clip]{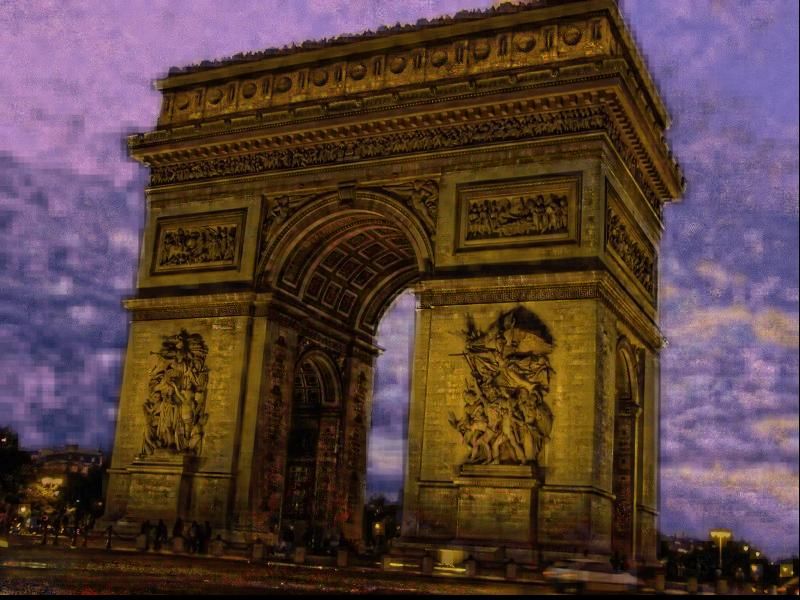}
			}
			\hspace{\wsp}
			\subfigure[]{
				\includegraphics[width=\width,clip]{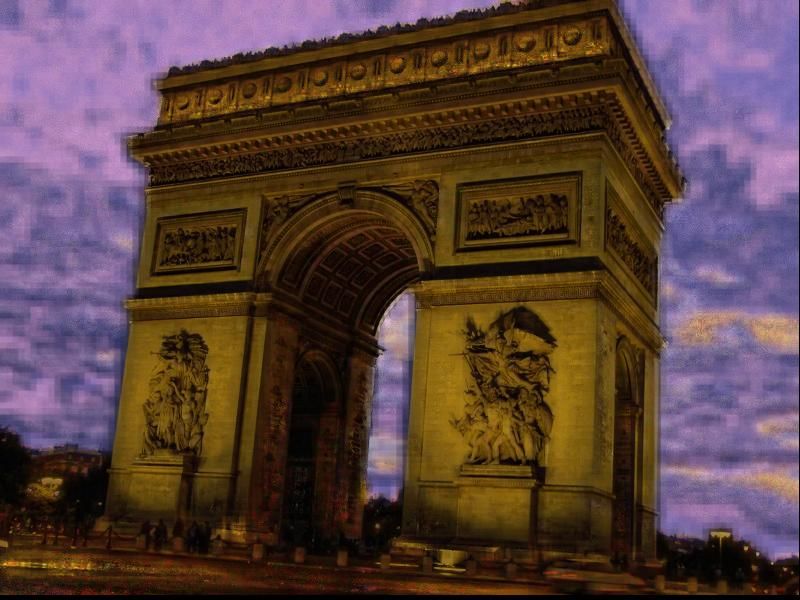}
			}
			\hspace{\wsp}
			\subfigure[]{
				\includegraphics[width=\width,clip]{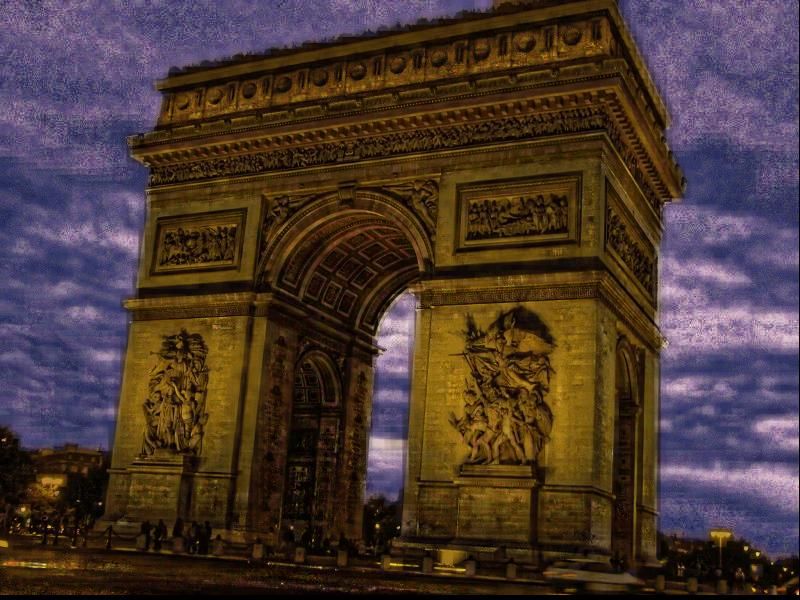}
			}
			\hspace{\wsp}
			\subfigure[]{
				\includegraphics[width=\width,clip]{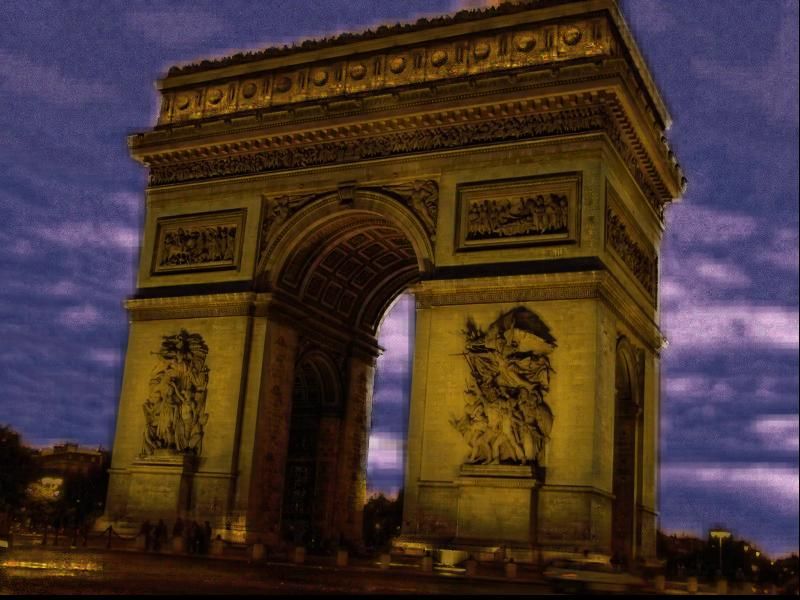}
			}	
			\hspace{\wsp}
			\subfigure[]{
				\includegraphics[width=\width,clip]{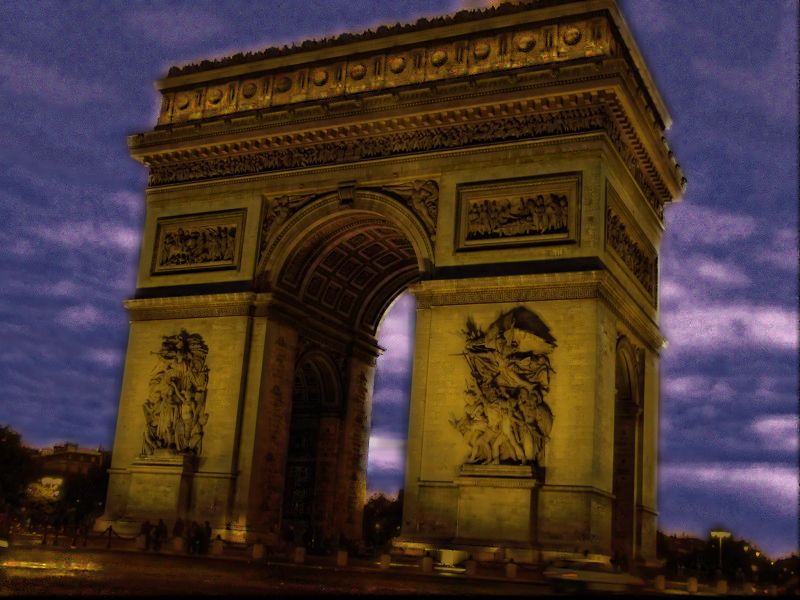}
			}																											
		}
	\end{center}
	\vspace{-8mm}
	
	\caption{The validation of the hierarchical features in SuperBIG. (a) Color + Distance. (b) Patch intensity vector. (c) Color + Distance + Texture. (d) Color + Distance + Texture + Patch intensity vector. (e) Color + Distance + Texture + Patch intensity vector + Gradient. }
	\label{fig:featureall}
\end{figure*}

\section{Conclusion and Future Works}
\label{sec:conclusion}
In this paper, we first introduce the concept of image stylistic brush and accordingly design an exemplar-based photo stylization method, SuperBIG, powered by a two-step bipartite graph algorithm. Specifically, a bipartite graph is constructed by considering dense correspondence and hierarchical features to partition pixels of the input and reference images into superpixels first. Then, we generate a superpixel-level bipartite graph, which produces correspondences of the superpixels by bipartite matching. The correspondence is then used to guide the style transformation in a decorrelated color space. Extensive experimental results demonstrate that the proposed SuperBIG method achieves superior visual quality compared to state-of-the-art methods while providing style consistent with the reference image.

Although SuperBIG shows very promising results in the extensive experiments, there is still room for improvement. First, SuperBIG assumes that the input and reference images contain the same scene. How to utilize the general category information to enable a more general exemplar-based stylization is worth exploring. Second, due to the graph structure of SuperBIG, it is time-consuming and difficult to apply in real applications. Thus, we aim to explore ways to speed up the processing with some optimizations~(such as image rescaling), in order to facilitate real-world applications.

\vspace{-1mm}
\bibliographystyle{abbrv}
\bibliography{sigproc}
\end{document}